\documentclass[10pt,twocolumn,letterpaper]{article}

\usepackage[pagenumbers]{cvpr} % To force page numbers, e.g. for an arXiv version
\usepackage{graphicx}
\usepackage{amsmath}
\usepackage{amssymb}
\usepackage{booktabs}
\usepackage{multicol,multirow}
\usepackage{makecell}
\usepackage[table]{xcolor}
\usepackage{xurl}

\usepackage[pagebackref,breaklinks,colorlinks]{hyperref}
\usepackage{colortbl}
\definecolor{mygray}{rgb}{0.89, 0.93, 0.85}
\definecolor{whitesmoke}{rgb}{0.96, 0.96, 0.96}
\definecolor{timberwolf}{rgb}{0.86, 0.84, 0.82}

\usepackage[capitalize]{cleveref}
\crefname{section}{Sec.}{Secs.}
\Crefname{section}{Section}{Sections}
\Crefname{table}{Table}{Tables}
\crefname{table}{Tab.}{Tabs.}

\usepackage{xspace}

\newcommand{\ours}{MobileVLM\xspace}
\newcommand{\ourllm}{MobileLLaMA\xspace}
\newcommand{\lammacpp}{\texttt{llama.cpp} }
\newcommand{\lammacppn}{\texttt{llama.cpp}}
\newcommand{\qks}{\texttt{q4\_k\_s} }
\newcommand{\qksn}{\texttt{q4\_k\_s}}
\newcommand{\qeight}{\texttt{q8\_0} }

\def\cvprPaperID{*****} % *** Enter the CVPR Paper ID here
\def\confName{CVPR}
\def\confYear{2022}

\begin{document}

%\title{\ours: A Fast, Reproducible and Strong Vision Language Assistant for Mobile Devices}
% CS: For discussion
\title{\ours: A Fast,  Strong and Open Vision Language Assistant \\ for Mobile Devices}

\author{Xiangxiang Chu$^{1}$, ~~ Limeng Qiao$^{1}$, ~~ Xinyang Lin$^{1}$, ~~ Shuang Xu$^{1}$, ~~ Yang Yang$^{1,3}$\thanks{Work done as an intern at Meituan.}, ~~ Yiming Hu$^{1}$,\\
Fei Wei$^{1}$, ~~ Xinyu Zhang$^{1}$, ~~Bo Zhang$^{1}$, ~~Xiaolin Wei$^{1}$, ~~Chunhua Shen$^{2}$\\[.2cm]
{\normalsize $^{1}$ Meituan Inc.}~~~~
{\normalsize $^{2}$ Zhejiang University, China}
~~~~
{\normalsize $^{3}$ Dalian University of Technology, China}
\\
% \and
}
\maketitle

%%%%%%%%% ABSTRACT
\begin{abstract}
   We 
   %inaugurate
   present 
   \ours, a competent multimodal vision language model (MMVLM) targeted to run on mobile devices. It is an amalgamation of a myriad of architectural designs and techniques that are mobile-oriented, which comprises a set of language models at 
   %a
   the 
   scale of 1.4B and 2.7B parameters, trained from scratch, a multimodal vision model that is pre-trained in the CLIP fashion, cross-modality interaction via an efficient projector. We evaluate \ours on several typical VLM benchmarks.
   %to have
    Our models demonstrate 
   % comparable
   \textit{on par} 
   performance compared  
   with a few much larger models. More importantly, we measure the inference speed on both a Qualcomm Snapdragon 888 CPU and an NVIDIA Jeston Orin GPU, %where
   and 
   we obtain 
   %a
   state-of-the-art performance 
   %at
   of 
   % 
   %21.54 tokens/s and 65.27 tokens/s 
   %
   21.5 tokens and 65.3 tokens per second,
   respectively. Our models
   %will be made available 
   are available 
   at: \url{https://github.com/Meituan-AutoML/MobileVLM}
\end{abstract}

%%%%%%%%% BODY TEXT
\section{Introduction}
\label{sec:intro}

Large multimodal models (LMMs), especially the family of Visual Language Models (VLMs), rise as a promising research direction for building general-purpose assistants due to their substantially enhanced capability in both perception and reasoning  \cite{yin2023survey}. However, it has been challenging to connect the representations of the pre-trained large language models (LLMs)   \cite{hoffmann2022training,zhang2022opt,chung2022scaling,vicuna,touvron2023llama} and the vision models to unveil the cross-modality properties, such as visual question answering, image captioning, visual knowledge reasoning and conversation, \textit{etc.}

Remarkable performance on this task can be witnessed in GPT-4V  \cite{gpt4v} and Gemini  \cite{gemini}, and the evaluation of their abilities can be found in \cite{akter2023depth,fu2023gemini}. However, %scarce
very limited 
technical details are available for these proprietary models. Simultaneously in the research community, a line of language tuning methods have been proposed  \cite{alayrac2022flamingo,li2023blip2,dai2023instructblip,liu2023llava,liu2023improved,liu2023llavaplus,Qwen-VL,zhu2023minigpt,ye2023mplug,wei2023lenna}. For instance, Flamingo  \cite{alayrac2022flamingo} exploits visual tokens to condition the frozen language model via gated cross-attention layers. 
BLIP-2  \cite{li2023blip2}  argues that such interaction is insufficient and introduces a lightweight querying transformer (called Q-Former) that extracts the most useful features from the frozen vision encoder and feeds them directly into the frozen LLM. 
%QFormer is trained in a two-stage strategy to bootstrap vision and language representation step-by-step. 
MiniGPT-4  \cite{zhu2023minigpt} aligns a frozen visual
encoder from BLIP-2  \cite{li2023blip2} with a frozen language model Vicuna  \cite{vicuna} via only one projection layer.  Independently, LLaVA  \cite{liu2023llava} applies a simple trainable projector that converts the vision features into embedding tokens, which have the same dimension as the word embeddings to be processed by the language model altogether. 

Noticeably, training strategies also exhibit a shift to accommodate the large-scale multimodal data of great diversity. LLaVA %is
may be 
the first attempt to replicate the \emph{instruction tuning paradigm} from LLMs to the multimodal 
%field.
scenario. 
To generate multimodal instruction-following data, it feeds textual information %like
such as 
captions and bounding-box coordinates of images to language-only GPT-4  \cite{gpt4}. MiniGPT-4  \cite{zhu2023minigpt} is firstly trained on a combined image captioning dataset and then fine-tuned on a curated alignment dataset of image-text pairs. InstructBLIP \cite{dai2023instructblip}  enforces vision-language instruction tuning based on the pre-trained BLIP-2 model, where the Q-Former is trained on a diverse set of datasets organized in an instruction-tuning format. mPLUG-Owl  \cite{ye2023mplug} introduces a two-stage training strategy where the visual part is pre-trained first and the large language model LLaMA  \cite{touvron2023llama} is then fine-tuned with LoRA  \cite{hu2021lora} with instruction data from various sources.

%Albeit
Despite 
the advances mentioned above of VLMs, there is a natural demand to enable cross-modality capacities in resource-constrained scenarios. Gemini  \cite{gemini} surpasses state-of-the-art performance on a range of multimodal benchmarks and introduces mobile-scale VLMs with 1.8B and 3.25B parameters for low-memory devices. Common compression techniques %like
such as 
distillation and quantization are also exploited for this purpose. %This paper aims
\textit{We aim 
to build the first %reproducible
open, 
mobile-scale VLMs %with
trained using 
public datasets and available techniques to achieve visual perception and reasoning, %in
customized for 
resource-constrained %environments.
platforms. 
} Our contributions are as follows:
\begin{enumerate}
\item We present \ours, a full-stack remake of multimodal visual language models tailored for mobile scenarios. To our knowledge, we are the first to provide a detailed, reproducible, and strong vision language model from scratch. With controlled and open-source datasets, we build a set of high-performance foundation language models and multimodal models.

\item We make extensive ablation studies on the design of visual encoders and systematically evaluate the VLM performance sensitivity on 
%different
various 
training paradigms, input resolutions, and model sizes. 

\item We design an efficient projector between visual and text features, which better aligns multimodal features while reducing the inference budget.

\item Our model is crafted to run efficiently on mobile, %and IoT
low-power 
devices, with a measured speed of 21.5 tokens/s on a Qualcomm mobile CPU and 65.3 tokens/s on a Jeston Orin GPU respectively.

\item Our models perform comparably on a large body of VLM benchmarks, attesting their potential in numerous tasks in practice. Although we mainly focus on edge scenarios, our model outperforms many recent VLMs, which can only be supported by powerful GPUs in the cloud.
\end{enumerate}

\section{Related Work}
\label{sec:related}

\subsection{Vision Transformer} 
Vision Transformer  \cite{dosovitskiy2020image} is now the dominant backbone for visual perception. Its %follow-ups %like
sequel methods 
such as 
Swin  \cite{liu2021swin}, DeiT  \cite{pmlr-v139-touvron21a}, PVT  \cite{wang2021pyramid}, and Twins  \cite{chu2021Twins} have upgraded its original architectural design to strengthen its representational power and efficiency. The pre-training paradigm has also experienced several shifts, from image-supervised learning (\ie labeled classification) to unsupervised learning like masked auto-encoders  \cite{he2022masked}, and most recently to language-supervised training as advertised by CLIP~\cite{radford2021learning}, which empowered ViTs with unprecedented zero-shot capability. VLMo~\cite{bao2022vlmo} enhances CLIP with unified multimodality training.

\subsection{LLMs}
Large language models 
often 
come with billions of parameters and are pre-trained on extremely extensive text corpora, exhibiting emergent capabilities  \cite{wei2022emergent} that have not been witnessed before. They have reshaped the field of natural language processing and are being used in a wide range of applications. To date, proprietary LLMs like GPT-4  \cite{gpt4} prevail over open-sourced models. Nevertheless, the community is exuberant with the continuous model releases, %like
including 
GLM \cite{du2021glm}, BLOOM  \cite{laurencconbigscience}, OPT  \cite{zhang2022opt} and LLaMA series  \cite{touvron2023llama,touvron2023llama2}. Many %follow-up 
recent 
works  \cite{zheng2023judging,qwen} have been built on top of them.

Noticeably, there is a %potent
trend to build smaller language models, \ie, whose parameters are around 1B or fewer. To name a few, GPT-Neo  \cite{gpt-neo}, Pythia  \cite{biderman2023pythia}, GALACTICA\cite{GALACTICA}, OpenLLaMA  \cite{openlm2023openllama}, Phi  \cite{gunasekar2023textbooks,li2023textbooks}, Qwen  \cite{qwen} all ship language models at such sizes. Although privately collected high-quality data can significantly boost the performance of LLMs \cite{li2023textbooks,qwen},  our target is to build reproducible and efficient models, hence we do not utilize any non-public data for our research.

\subsection{VLMs}

\begin{table*}[ht]
  \centering
  \renewcommand{\arraystretch}{1.2}
  \setlength{\tabcolsep}{2pt}
  \scalebox{0.78}{
  \begin{tabular}{lllp{3.6cm}p{9cm}}
    \toprule
    Model & Vision Encoder & Language Model & Cross-modality Design & Multimodal Training Corpora \\
    \midrule
    CLIP  \cite{radford2021learning} & ViT, ResNet & Transformer & Linear-Projection & WebImageText~\cite{radford2021learning} (400M image-text pairs) \\
    BLIP  \cite{li2022blip} & ViT & MED$^*$ & Cross-Attention & 
    COCO \cite{lin2014microsoft}, VG  \cite{krishna2017visualgenome}, CC3M~\cite{sharma2018conceptual}, CC12M \cite{changpinyo2021conceptual}, LAION~\cite{schuhmann2021laion}
    %\multirow{2}{*}{\makecell[l]{COCO \cite{lin2014microsoft}, VG  \cite{krishna2017visualgenome}, CC3M~\cite{sharma2018conceptual}, CC12M \cite{changpinyo2021conceptual}, LAION~\cite{schuhmann2021laion} \\ (129M) }
    \\
    BLIP-2  \cite{li2023blip2} & CLIP/Eva-CLIP ViT & OPT, Flan-T5  & Q-Former & same as BLIP \\
    InstructBLIP  \cite{dai2023instructblip} & ViT-G/14 & Flan-T5, Vicuna & Q-Former w/ FC & 13 held-in out of 26 public datasets 
    % (See %its Appendix C)
    \\
    Flamingo  \cite{alayrac2022flamingo} & NFNet-F6  \cite{jiang2021all} & Chinchilla & Perceiver-Resampler
    % , ga\-ted c\-ro\-ss-att\-en\-ti\-on
    \cite{jaegle2021perceiver}
    & M3W(43M), ALIGN(1.4B) \cite{jia2021scaling}, LTIP (312M), VTP (27M) \\
    LLaVA  \cite{liu2023llava} & CLIP ViT-L/14 & Vicuna 7B/13B & Linear-Projection & filtered CC-595K from CC3M~\cite{sharma2018conceptual}, LLaVA-Instruct-158K \\
    % (595K image-text pairs) for pre-training, 158K for instruction tuning
    % 595K for pre-training, 158K for instruction-tuning \\
    LLaVA-1.5  \cite{liu2023improved} & CLIP ViT-L@336 & Vicuna-7B/13B & 
    %2-layer
    MLP  & a subset of InstructBLIP (1.2M)  \\
    MiniGPT-4  \cite{zhu2023minigpt} & EVA-CLIP ViT-G/14 & Vicuna-7B & Q-Former & LAION, CC, SBU  \cite{ordonez2011im2text} \\
    Shikra  \cite{chen2023shikra} & CLIP ViT-L/14 & Vicuna-7/13B & 
    %1-layer 
    FC-layer & Flickr30K  \cite{plummer2015flickr30ke}, RefCOCO  \cite{kazemzadeh2014refcoco}, VG, Visual-7W  \cite{mani2020pointqa} \\
    mPLUG-Owl \cite{ye2023mplug} &  CLIP ViT-L/14 & LLaMA-7B  & Cross-Attention
    %, LoRA instruction tuning 
    & LAION-400M, COYO-700M  \cite{kakaobrain2022coyo-700m}, CC, COCO\\
    %KOSMOS-1  \cite{huang2023language} & VLMo  & MAGNETO  \cite{wang2022foundation} & Perceiver resampler & LAION-2B, LAION-400M, COYO-700M, CC \\
    KOSMOS-2  \cite{peng2023kosmos}  & VLMo  \cite{bao2022vlmo}  & MAGNETO  \cite{wang2022foundation} & Perceiver-Resampler & GRIT (curated with LAION-2B, COYO-700M) \\
    QWen-VL   \cite{Qwen-VL} & Openclip ViT-bigG  \cite{openclip} &  Qwen-LM & Cross-Attention & multi-tasking datasets (Captioning, VQA, Grounding, etc.
    %, set its see Table 3
    )\\
    ShareGPT4V  \cite{chen2023sharegpt4v}  & CLIP ViT-L/14@336 & Vicuna-7B & 
    %2-layer
    MLP &  ShareGPT4V (100K by GPT-4V, 1.2M by its learned
    % caption 
    model) \\
    %PandaGPT & &  & \\
    \rowcolor{mygray} \ours (ours)  & CLIP ViT-L/14@336 & \ourllm & LDP & same as LLaVA-1.5 \cite{liu2023improved} \\
    \bottomrule
  \end{tabular}
  }
  \caption{Comparison of open-source VLM architectures and their training corpora. $^*$: BLIP adopts a multimodal encoder/decoder.}
  \label{tab:VLMs}
\end{table*}

Throughout recent years, a school of vision language models has rapidly emerged. 
  Table~\ref{tab:VLMs} summarizes them in a detailed comparison regarding architectures, cross-modality design, and training corpora. 
%   CLIP  \cite{radford2021learning}, BLIP-2  \cite{li2023blip2}, Flamingo  \cite{alayrac2022flamingo},  Perceiver\cite{jaegle2021perceiver} Resampler Q-Former GPT4v ShareGPT4V  \cite{chen2023sharegpt4v} , QWen-VL  \cite{Qwen-VL}, ChatGPT LLaVA  \cite{liu2023llava} LLaVA-v1.5, MiniGPT-4  \cite{zhu2023minigpt}
% mPLUG-Owl  \cite{ye2023mplug} KOSMOS-2  \cite{peng2023kosmos} Shikra  \cite{chen2023shikra}
%CLIP  \cite{radford2021learning} is the pioneering work that collects massive image-text pairs and adopts image-text contrastive training to achieve high-quality alignment of image-text models. BLIP-2  \cite{li2023blip2} uses a lightweight Q-former structure to efficiently complete modal alignment on frozen visual and text pretraining models. Flamingo  \cite{alayrac2022flamingo} exploits Perceiver\cite{jaegle2021perceiver} and gated cross-attention to fuse multimodal information and demonstrates strong few-shot capabilities. LLaVA  \cite{liu2023llava} adopts linear layers as projectors to feed visual features to LLMs and enhances the instruction following ability of visual multimodal models through visual instruction learning. ShareGPT4V  \cite{chen2023sharegpt4v} uses GPT4-vision  \cite{gpt4v} to build a high-quality image-text caption dataset, and it has been proven in both the pre-train and Supervised Fine-Tuning (SFT) stages that high-quality caption datasets can effectively enhance the capabilities of VLM models.

\textbf{Architecture choices.}
As a consequence of the intimidating training cost of large language models, most language models used in VLMs are pre-trained open-source models like OPT  \cite{zhang2022opt}, Flan-T5  \cite{chung2022scaling}, Chinchilla  \cite{kudo2018sentencepiece}, Vicuna  \cite{vicuna} and LLaMA  \cite{touvron2023llama}. QWen adapts LLaMA with custom variations  \cite{qwen} to obtain an LLM on their own. 

Visual backbones in VLMs are typically vision transformer \cite{dosovitskiy2020image}, but pre-trained in various strategies  \cite{radford2021learning,li2023blip2,fang2023eva}. Most VLMs prefer CLIP-fashioned ViT  \cite{radford2021learning} trained with natural language supervision. Flamingo picks NFNet-F6  \cite{jiang2021all}.  KOSMOS chooses VLMo  \cite{bao2022vlmo} instead.

\textbf{Dataset-centric-ness.}
The construction of training data has become increasingly crucial. It is common to utilize millions of text-image pairs in the line of VLMs, where the new datasets are usually released alongside their corresponding new models. To name a few, 
apart from an enhanced visual receptor and novel language model called Qwen-LM  \cite{qwen}, the multilingual multimodal Qwen-VL  \cite{Qwen-VL} additionally aligns the image with caption and box tuples, which sets a new record of generalist models. 
PALI  \cite{chen2022pali} and PALI-X  \cite{chen2023palix} consume an internal multi-language image-text dataset called WebLI at a scale of 12 billion.
Most recently, observing the constraints of current image-text datasets like hallucination and inaccurate descriptions, ShareGPT4V  \cite{chen2023sharegpt4v}  exploits GPT-4V  \cite{gpt4v} for generating 1.2M high-quality image-text pairs with which can surpass the LLaVA series. Similarly built with GPT-4V, LVIS-INSTRUCT4V  \cite{wang2023see} helps LLaVA-1.5 to gain substantial improvement on various VLM benchmarks.

\subsection{Model Compression for LLMs}
Large Language Models (LLMs) have brought 
%about
a paradigm shift in natural language processing, while their colossal size and computational requirements pose significant challenges for real-world deployment, particularly in environments with limited resources. The size of these models often results in high memory usage and slow processing speeds. Additionally, the energy requirements for training and operating these models raise sustainability concerns. These challenges are becoming more pressing as LLMs continue to grow in size and complexity. In response to these challenges, model compression has emerged as a crucial research area, which aims to reduce the size and computational demands of models without significantly compromising their performance. These techniques include but not limited to model pruning  \cite{ma2023llmpruner,frantar2023sparsegpt,sun2023simple}, quantization  \cite{frantar2022gptq,xiao2023smoothquant,li2023norm}, knowledge distillation  \cite{zhang2023lifting} and low-rank decomposition \cite{yao2023zeroquantv2}.

Besides, LLM deployment tools have experienced a blossom, evidenced by industrial frameworks like TensorRT-LLM  \cite{tensorrt-llm}, LMDeploy  \cite{lmdeploy}, and \lammacpp  \cite{Gerganov2019} %are
being 
developed to tackle deployment difficulties in diverse environments.

%\subsection{Downscaled VLMs}
%Kosmos-2   \cite{peng2023kosmos} 
%Phi  \cite{li2023textbooks} is a tiny 
%

%Apart from VLMs, there is a tendency to downscale LLMs to facilitate mobile uses, e.g. TinyLLaMA, Phi.

\subsection{VLM Benchmarks}
%Vision-language models have gained prominence as a pivotal research area and shown remarkable scene understanding, perception, and reasoning abilities. Meanwhile, 
Systematic and comprehensive evaluations of different VLMs are of great necessity. To this end, POPE   \cite{li2023evaluating} provides a benchmark for evaluating hallucination in VLMs, which formulates the evaluation as a binary classification task that prompts the VLMs to answer whether the object exists. In contrast, GQA   \cite{hudson2019gqa} mainly centers around the VLMs' abilities in real-world reasoning, scene understanding, and compositional question answering. TextVQA   \cite{singh2019towards} consists of questions related to the text in the image, evaluating the OCR and reasoning abilities of models. ScienceQA   \cite{lu2022learn} consists of multimodal multiple-choice questions covering scientific topics, \eg, natural science, social science, and language science, which requires VLMs to integrate common sense into reasoning. MME  \cite{fu2023mme} measures both the perception and cognition abilities of VLMs, it covers a total of 14 subtasks varying from coarse-grained to fine-grained ones.
MMBench  \cite{liu2023mmbench} is a methodically constructed multimodal dataset, which collectively covers a diverse spectrum of 20 fine-grained skills and involves a circular evaluation strategy with the incorporation of ChatGPT  \cite{ChatGPT}.

\subsection{Embodied AI} 
Our work is closely related to Embodied Artificial Intelligence. Being one of the central goals of Artificial General Intelligence, embodied AI strives to build egocentric intelligence systems that can interact with their surroundings with perception, reasoning, and planning capabilities   \cite{duan2022survey}. Recently, the emerging large vision language models   \cite{mu2023embodiedgpt,song2023llm,sumers2023distilling} allow embodied AI agents to resolve the relevant tasks like \emph{embodied question answering}  \cite{das2018embodied}, and \emph{vision-language navigation}  \cite{li2019robust,pashevich2021episodic} in a highly end-to-end fashion.  

\section{\ours}
\label{sec:method}

\subsection{Overall Architectural Design}

\begin{figure*}[t!]
  \centering
   \includegraphics[width=.7\linewidth]{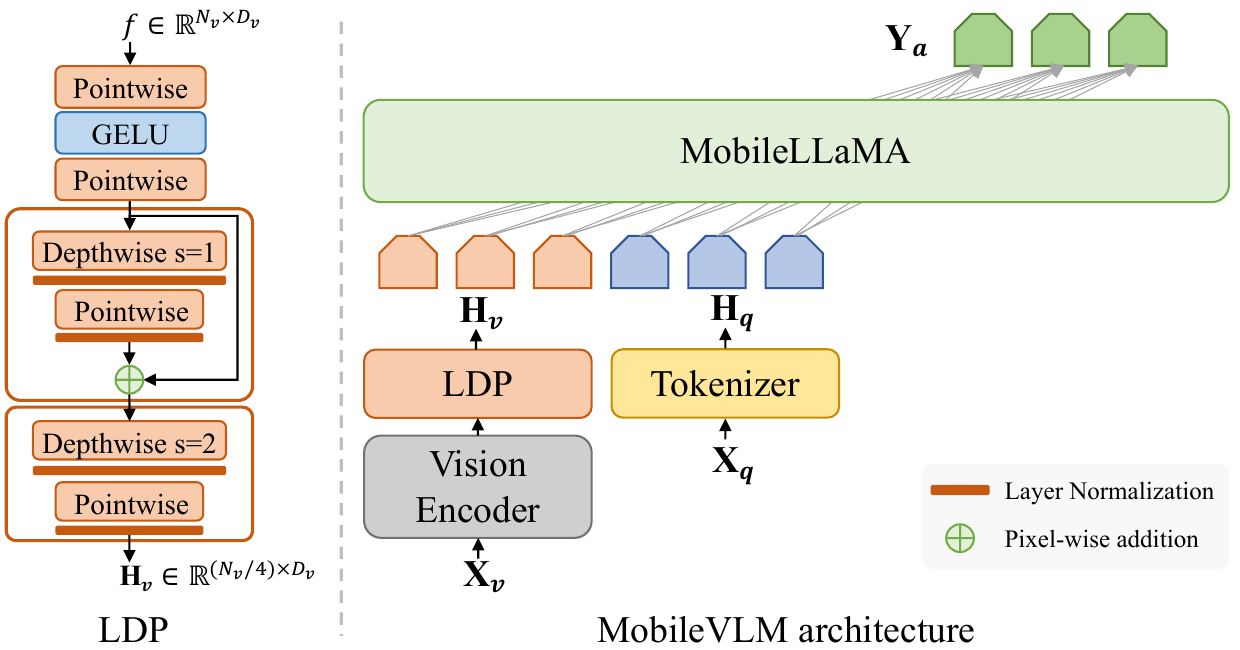}
   \caption{The \ours architecture (\textit{right}) utilizes \ourllm as its language model, intakes $\mathbf{X}_v$ and $\mathbf{X}_q$ which are image and language instructions as respective inputs and gives $\mathbf{Y}_a$ as the output language response. LDP refers to a lightweight downsample projector (\textit{left}).}
   \label{fig:mobile-llava-arch}
\end{figure*}

With the primary goal of achieving efficient visual perception and reasoning for resource-limited edge devices in mind, we design the overall architecture of \ours as illustrated in Figure~\ref{fig:mobile-llava-arch}. It contains three components: \textit{1) a visual encoder, 2) a tailored LLM for edge devices (%codenamed 
\ourllm), and 3) an efficient projector (%called
termed ``lightweight downsample projector'', LDP) that %bridges the gap between
aligns 
the visual and the textual space. 
}

Taking an image $\mathbf{X}_v \in \mathbb{R}^{H\times W\times C}$ as input, the vision encoder $\mathbf{F}_{enc}$ extracts the visual embeddings $f\in \mathbb{R}^{N_{v}\times D_{v}}$ for image perception, where $N_{v}={HW/P^{2}}$ denotes the number of image patches and $D_v$ denotes the hidden size of visual embeddings. To alleviate the efficiency issues arising from prolonged image tokens, we scheme a lightweight projector $\mathbf{P}$ for visual feature compression and visual-text modal alignment. It converts ${f}$ into the word embedding space with an appropriate input dimension of the subsequent language model as below,
\begin{equation}
\mathbf{H}_{v}=\mathbf{P}({f}),  {f}=\mathbf{F}_{enc}(\mathbf{X}_v).
\end{equation}

Thus we obtain the image tokens $\mathbf{H}_{v}\in \mathbb{R}^{(N_{v}/4)\times D_{t}}$ and text tokens $\mathbf{H}_{q}\in \mathbb{R}^{N_{t}\times D_{t}}$, where $N_{t}$ denotes the number of text tokens and $D_{t}$ denotes the hidden size of the word embedding space. Observing that LLM occupies the most computation and memory consumption in the current design paradigm of MLLMs, we tailor a series of inference-friendly LLMs that enjoy advantages in speed for mobile applications. It predicts the response $\mathbf{Y}_{a}=\left \{ y_{i} \right \} ^{L}_{i=1}$ conditioned on the multimodal input in an autoregressive manner, where $L$ denotes the output token length. This process can be formulated as,
\begin{equation}
p({\mathbf{Y}_{a}}|{\mathbf{H}_{v}},{\mathbf{H}_{q}})=\prod_{i=1}^{L}p(y_{i}|{\mathbf{H}_{v}},{\mathbf{H}_{q}},y_{<i}).
\end{equation}

\subsection{Visual Encoder} % {\color{red}@qiaolimeng}}
Based on empirical analysis later shown in Sec~\ref{sec:ablation-vision}, we leverage the pre-trained CLIP ViT-L/14  \cite{radford2021learning} with a resolution of 336$\times$336 as our visual encoder $\mathbf{F}_{enc}$. The Vision Transformer (ViT)  \cite{dosovitskiy2020image} dissects images into uniform-sized patches, applies a linear embedding to each, and integrates positional encodings before feeding the resultant vector sequence into a canonical Transformer encoder. Typically, a classification token is appended to this sequence for subsequent categorization tasks. 

% CLIP leverages contrastive learning to pre-train on a substantial dataset comprising image-text pairs, culminating in an image encoder (ViT) that furnishes a nuanced visual representation, which exhibits a strong capacity for generalization and adaptability across a variety of tasks. 

\subsection{\ourllm}
For the language model, we downscale LLaMA \cite{touvron2023llama} to facilitate the off-the-shelf deployment, \ie, our models can be seamlessly supported by almost all popular inference frameworks. Moreover, we evaluate the model latency on the edge devices to guide the model design. Neural architecture search \cite{liu2018darts,chu2021fairnas, howard2019searching, chu2020fair, cai2020once} would be a better choice, but for the time being we keep it as our future work. The detailed setting of our architecture is shown in Table~\ref{tab: LLM_setting}. 

\begin{table}[ht]
  \centering
  \setlength{\tabcolsep}{2pt}
  \begin{tabular}{@{}lccccc@{}}
    \toprule
    Model & Blocks & Dim&  Heads & Context length\\
    \midrule
    \ourllm 1.4B & 24&2048& 16 & 2k\\
    \ourllm 2.7B & 32 &2560&32 & 2k\\
    \bottomrule
  \end{tabular}
  \caption{Detailed settings of our language models.}
  \label{tab: LLM_setting}
\end{table}

Specifically, we utilize the sentence piece tokenizer \cite{kudo2018sentencepiece} from LLaMA2 \cite{touvron2023llama2} with a vocabulary size of 32000 and train the embeddings from scratch. This is beneficial for performing future distillation without further pain. The context length used at the pre-train stage is 2k for all models due to limited resources. However, the context window can be further scaled to 8k for inference, as indicated by \cite{chen2023extending}. The detailed settings of other components are listed below.
\begin{itemize}
    \item We apply RoPE \cite{su2023roformer} to inject positional information.
    \item We apply pre-normalization to stabilize training. Specifically, we use RMSNorm \cite{zhang2019root} instead of layer norm and the MLP expansion ratio 8/3 instead of 4. 
    \item We also use SwiGLU activation function \cite{shazeer2020glu} instead of GELU as \cite{touvron2023llama}.

\end{itemize}

\subsection{Efficient Projector} 
\label{Efficient Projector}
The projector between the vision encoder and the language model is critical in aligning multimodal features. There are two existing paradigms: Q-Former  \cite{zhu2023minigpt,li2023blip2} and MLP projection  \cite{liu2023llava,liu2023improved}. Q-Former explicitly controls the number of visual tokens per query to force extracting the most relevant visual information. However, it inevitably loses the spatial positional information of tokens and suffers from slow convergence. In addition, it is  %not
inefficient for the inference on edge devices. In contrast, MLP retains the spatial information but it usually includes useless tokens such as the background. For an image of $\mathbf{X}_v \in \mathbb{R}^{H \times W\times C}$ with a patch size of $P$, there are $N_{v}={HW/P^{2}}$  visual tokens to be injected into the LLM model, which greatly slows down the overall inference speed. Inspired by \cite{chu2023conditional}, we can utilize convolutions to enhance the positional information and encourage local interaction of the vision encoder. Specifically, we explore mobile-friendly operations based on depth-wise convolution (the simplest form of PEG  \cite{chu2023conditional}), which is efficient and well-supported by various edge devices. 

To keep spatial information and to minimize the computational cost, we make use of a convolution with a stride of 2, which reduces 75\% visual tokens.  This design significantly boosts the overall inference speed. However, our experimental result indicates that solely downsampling the tokens severely deteriorates the performance on downstream tasks, such as OCR. To alleviate this effect, we utilize a more powerful network instead of a single PEG. The detailed architecture of the efficient projector, called Lightweight Downsample Projector (LDP), is illustrated in Figure~\ref{fig:vlm-training-strategy}. Note that, this projector network only contains less than 20M parameters and runs about $81 \times$ faster than the visual encoder. %In other words, its contribution to the whole network latency can be neglected, for which we name it efficient projector.

% \begin{figure}[ht]
%   \centering
%    \includegraphics[width=\linewidth]{figures/token_vs_speed.pdf}
%    \caption{Speed of \ours w.r.t. the number of input tokens per visual prompt.}
%    \label{fig:token-speed}
% \end{figure}

We use Layer Normalization instead of Batch Normalization \cite{ioffe2015batch} to make training stable and not affected by the batch size. Since the projector is already very light-weight, therefore, we don't adopt recent mobile reparameterization designs \cite{vasu2023mobileone,chu2023make}.

Formally, LDP (denoted as $\mathbf{P}$) takes the
visual embeddings $f\in \mathbb{R}^{N_{v}\times D_{v}}$ as input, and outputs the efficiently extracted and aligned visual tokens $\mathbf{H}_{v}\in \mathbb{R}^{(N_{v}/4)\times D_{t}}$ as,
\begin{equation}
\mathbf{H}_{v}=\mathbf{P}(f)= \begin{cases}
  f_{0}&=PW(GELU(PW(f))), \\  
  f_{1}&=LN(PW(LN(DW(f_{0}))))+f_{0}, \\
\mathbf{H}_{v}&=LN(PW(LN(DW(f_{1}))).
\end{cases}
\end{equation}

\begin{table*}[ht]
  \centering
  \setlength{\tabcolsep}{3pt}
  \begin{tabular}{l*{12}{c}}
    \toprule
    % common sense reasoning tasks language understanding tasks
    \multirow{2}{*}{\textbf{Model}} & \multicolumn{5}{c}{Common Sense Reasoning} && \multicolumn{3}{c}{Language Understanding} \\\cline{2-6} \cline{8-10}
    & ARC$_{(c/e)}$ & BoolQ & RTE &  Winogrande & TruthfulQA && HellaSwag & PIQA & MMLU & Avg. \\
    \midrule
    INCITE 3B (V1) \cite{together2023redpajama} & 0.32 / 0.68 & 0.67 &0.52& 0.63 & 0.21 / 0.33 &&0.48&  0.74 & 0.2675 & 0.4848 \\
    % BLOOM 3B & & & & & & & & & &&\\
    OpenLLaMA 3B (V1) \cite{openlm2023openllama} & 0.34 / 0.69 & 0.68 & 0.58 & 0.62 & 0.22 / 0.35 && 0.49 &  0.75& 0.2695 & 0.4990 \\
    \rowcolor{mygray} \ourllm 2.7B & 0.32 / 0.68& 0.61& 0.59 & 0.63 & 0.23 / 0.36&&0.48& 0.75& 0.2730 & 0.4923 \\
    \midrule
    % BLOOM 1.1B & & & & & & & & & &&\\
    TinyLLaMA 1.1B (2T) \cite{tinyllama}& 0.23 / 0.57& 0.59 &0.55 & 0.57 & 0.23 / 0.39&&0.40& 0.70 & 0.2541& 0.4484 \\
    Galactica 1.3B \cite{GALACTICA} & 0.28 / 0.62 & 0.62 & 0.52  & 0.55& 0.25 / 0.41&&0.34& 0.63&0.2675 & 0.4488\\
    OPT 1.3B \cite{zhang2022opt} & 0.24 / 0.57 & 0.56 & 0.51 & 0.59 & 0.24 / 0.39 &&0.41 & 0.71 & 0.2461 &0.4466 \\
    Pythia 1.4B \cite{biderman2023pythia} & 0.26 / 0.61 & 0.63 &0.52  & 0.57 & 0.23 / 0.39 &&0.40& 0.71 & 0.2568 & 0.4577\\
    % GPT-neo 1.3B & & & & & & & & &  &&\\
    \rowcolor{mygray}\ourllm 1.4B & 0.26 / 0.61& 0.53& 0.53& 0.59 & 0.21 / 0.35&&0.43& 0.73& 0.2497 &0.4490 \\
    \bottomrule
  \end{tabular}
  \caption{Comparison with SOTA mobile-scale language models on mainstream language benchmarks.}
  \label{tab:llm-comparison}
\end{table*}

\section{Experiment}
\label{sec:exp}

\subsection{Training}
The whole reproducible training process is composed of three stages. Firstly, we pre-train LLM foundation models on the text-only dataset RedPajama v1  \cite{together2023redpajama}. Secondly, we perform supervised fine-tuning (SFT) following Vicuna  \cite{vicuna2023} on a dataset of multi-turn dialogues between humans and ChatGPT from third-party platforms. Lastly, we train our vision large models using multimodal datasets.

\begin{figure}[t]
  \centering
   \includegraphics[width=\linewidth]{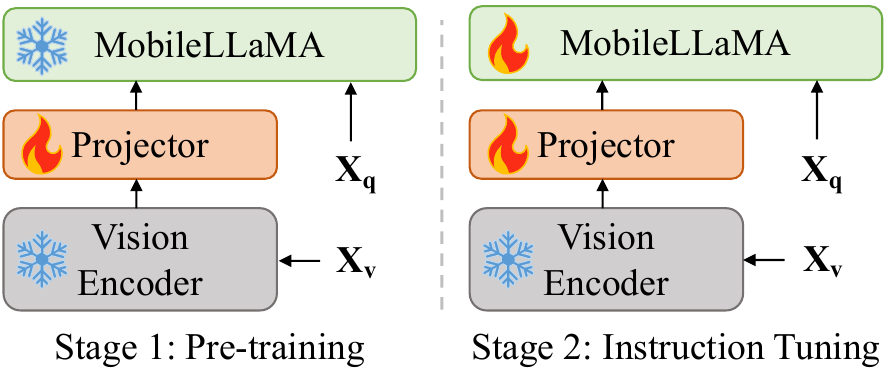}
   \caption{Illustration of the \ours training strategy.}
   \label{fig:vlm-training-strategy}
\end{figure}

\textbf{Language model pre-training.} 
Since our target is training deployable models for inference, we do not strictly follow the efficient combination of model capacity and tokens from scaling law \cite{kaplan2020scaling}.  \textit{To make our work reproducible, all the models are trained on 1.3T tokens\footnote{Applying the tokenizer of \cite{touvron2023llama2} on this dataset generates about 1.3T tokens.} from the RedPajama v1 \cite{together2023redpajama} dataset only}. This benefits further research by enabling controlled experiments. We apply the same sampling ratio of different datasets as \cite{touvron2023llama}, which is shown in Table~\ref{tab: LLM_traing_data_proportion} (see Appendix). The common autoregressive loss is adopted. We utilize a global batch size of 5,242,880. 

The peak learning rate is set to 3 $\times$ 10$^{-4}$ and it decays to 3 $\times$ 10$^{-5}$ following the cosine strategy. We warm up with 2000 iterations.  We use the AdamW optimizer \cite{loshchilov2017decoupled} with $\beta_1=0.9$ and $\beta_2=0.95$ and a weight decay regularization value of 0.1. The gradient clipping threshold is set to 1.0. We adopt the Pytorch lightning framework with DeepSpeed backend \cite{rasley2020deepspeed}. Specifically, we utilize ZERO 1 \cite{rajbhandari2020zero} and gradient accumulation to achieve a training speed of 18800 TGS (Tokens per second for a GPU) for the 1.4B model and 8500 TGS for the 2.7B model, on 20 nodes equipped with 8 NVIDIA Tesla A100 GPUs each. 

\begin{figure}[t]
  \centering
   \includegraphics[width=\linewidth]{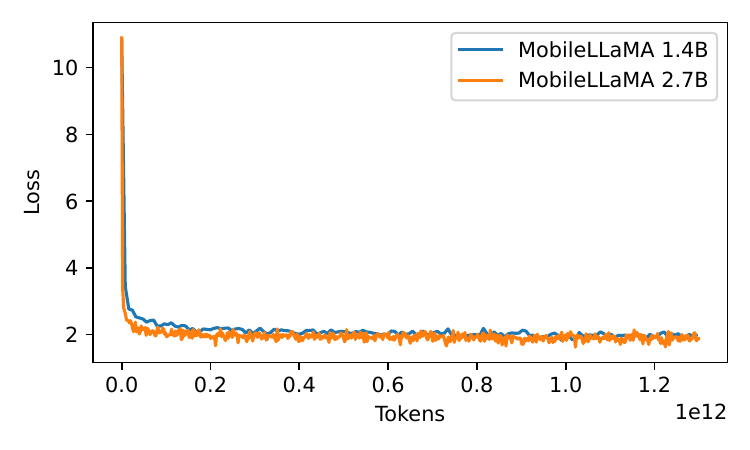}
   \caption{Training loss curves of our \ourllm{ 1.4B} and 2.7B on 1.3T tokens of RedPajama data.}
   \label{fig:curve}
\end{figure}

Furthermore, we also favor Flash Attention V2 \cite{dao2023flashattention} to alleviate the I/O bottleneck and to train faster. We randomly shuffle the data to disturb the sequential order with a fixed seed, which is vital since the training process can be intermittently interrupted and requires to be resumed. We first tokenize the raw data into IDs and serialize them into many bucket files. We then utilize memory mapping to deliver a desired I/O speed. Besides, we pack different sentences together, where an \texttt{EOS} token is inserted to set different sentences apart. Due to limited resources, we do not try %out
the design of InternLM  \cite{2023internlm}, which may further improve the model performance by disabling such packing. The overall training loss decreases as the consumed tokens increase and is shown in Figure~\ref{fig:curve}.

\begin{figure}[t]
  \centering
   \includegraphics[width=\linewidth]{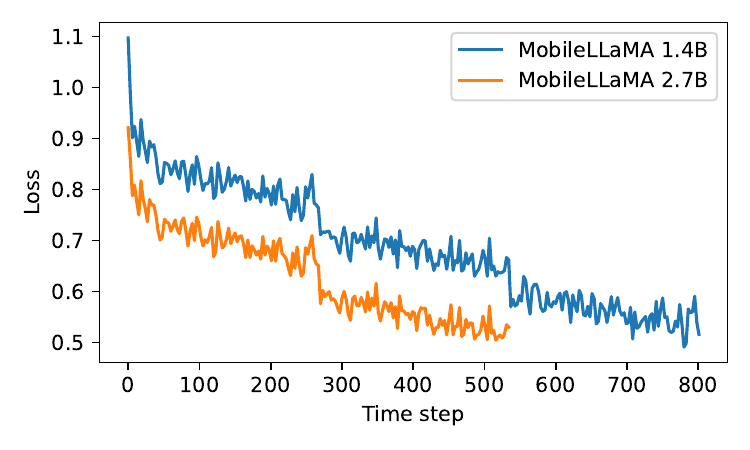}
   \caption{SFT loss curves of \ourllm 1.4B and 2.7B.}
   \label{fig:sft-curve}
\end{figure}

\textbf{SFT on language models.} %{\color{red}{@linxinyang, we need a training loss curve here}}{\color{green}{ Done, need review}}
As clarified by LLaMA-2  \cite{touvron2023llama2}, fewer higher-quality examples from their vendor-based annotation efforts significantly improve the results. We are thus motivated to finetune our \ourllm on a high-quality dataset with supervised learning. Vicuna \cite{vicuna2023} fine-tunes LLaMA on user-shared conversations collected from ShareGPT, which is widely used as a language module for multimodal model building, but their training dataset is not released. We employ a dataset of multi-turn dialogues between humans and ChatGPT from third-party platforms \cite{huggingface_data}
%\footnote{\url{https://huggingface.co/datasets/Aeala/ShareGPT_Vicuna_unfiltered}},
which has been cleaned through a process of format standardization and quality refinement. The SFT data is organized following the Vicuna format, where each sample consists of a dialogue including several user prompts and ChatGPT answers. As shown in Table~\ref{table:sft-prompt} (see Appendix), a special token is utilized to separate the assistant's answer and the next round of user prompts. For the training details, we use a cosine learning rate schedule without weight decay, a global batch size of 128, and a sequence length of 2048 tokens. We utilize an autoregressive objective and perform backpropagation only on answer tokens. To achieve better performance in downstream tasks, we conducted experiments to select the appropriate hyperparameters. We fine-tune for 3 epochs with a learning rate of 
%2e-5
$ 2 \times 10^{-5}$
for \ourllm 1.4B, and 2 epochs with a learning rate of %3e-5
$ 3 \times 10^{-5}$
for \ourllm 2.7B. The training loss decreases with iterations as shown in Figure~\ref{fig:sft-curve}. To be later shown in Sec.~\ref{sec:exp-sft}, our empirical performance on downstream tasks demonstrates that high-quality SFT data is essential to aligning LLMs with dialogue-style instructions.

%% MOVED HERE FOR TYPESETTING
\begin{table*}[t!]
\centering
\setlength{\tabcolsep}{3pt}
\begin{tabular}{*{1}{l}*{4}{l}|*{6}{cc}}
\toprule
Method & LLM & Res. & PT & IT & GQA & SQA$^\text{I}$ & VQA$^\text{T}$ & POPE & MME & MMB$^\text{dev}$ \\
\midrule
Openflamingo \cite{anas2023OpenFlamingo} & MPT-7B & 336 & 180M & - & -- & -- & 33.6 & -- & -- & 4.6 \\
BLIP-2 \cite{li2023blip2} & Vicuna-13B & 224 & 129M & - & 41.0 & 61.0 & 42.5 & 85.3 & 1293.8 & -- \\
MiniGPT-4 \cite{zhu2023minigpt} & Vicuna-7B & 224 & 5M & 5K & 32.2 & -- & -- & -- & 581.7 & 23.0\\
InstructBLIP \cite{dai2023instructblip} & Vicuna-7B & 224 & 129M & 1.2M & 49.2  & 60.5 & 50.1 & -- & -- & 36.0  \\
InstructBLIP \cite{dai2023instructblip}& Vicuna-13B & 224 & 129M & 1.2M & 49.5  & 63.1 & 50.7 & 78.9 & 1212.8 & --  \\
Shikra \cite{chen2023shikra}& Vicuna-13B & 224 & 600K & 5.5M & -- & -- & -- & -- & -- & 58.8 \\
mPLUG-Owl \cite{ye2023mplug} & LLaMA-7B & 224 & 2.1M & 102K & -- & -- & -- & -- & 967.3 & 49.4  \\
IDEFICS-9B \cite{laurenccon2023obelisc} & LLaMA-7B & 224 & 353M & 1M & 38.4 & -- & 25.9 & -- & -- & 48.2 \\
IDEFICS-80B \cite{laurenccon2023obelisc} & LLaMA-65B & 224 & 353M & 1M & 45.2 & -- & 30.9 & -- & -- & 54.5 \\
Qwen-VL \cite{Qwen-VL} & Qwen-7B & 448 & 1.4B & 50M & 59.3 & 67.1 & 63.8 & -- & 1487.6 & 38.2 \\
MiniGPT-v2 \cite{chen2023minigpt} & LLaMA-7B & 448 & 23M & 1M & 60.3 & -- & -- & -- & -- & 12.2\\
LLaVA-1.5 \cite{liu2023improved} & Vicuna-7B & 336 & 558K & 665K & 62.0 & 66.8 & 58.2 & 85.9 & 1510.7 & 64.3\\
% Gemini Nano 2  \cite{gemini} & Nano 2-3.25B & - & - & - & -- & -- & 65.9 & -- & -- & -- \\
\midrule
\rowcolor{mygray}
\ours 1.7B & \ourllm 1.4B & 336 & 558K & 665K & 56.1 & 54.7 & 41.5 & 84.5 & 1196.2 & 53.2 \\
\ours 1.7B w/ LoRA& \ourllm 1.4B & 336 & 558K & 665K & 57.0 & 53.1 & 42.3 & 86.0 & 1143.7 & 50.4 \\
\rowcolor{mygray}
\ours 3B & \ourllm 2.7B & 336 & 558K & 665K & 59.0 & 61.0 & 47.5 & 84.9 & 1288.9 & 59.6 \\
\ours 3B w/ LoRA & \ourllm 2.7B & 336 & 558K & 665K & 58.4 & 59.0 & 46.7 & 84.6 & 1296.4 & 57.0\\
\bottomrule
\end{tabular}
% \vspace{-1mm}
\caption{\textbf{Comparison with SOTA methods on 6 VLM benchmarks.} GQA  \cite{hudson2019gqa}; SQA$^\text{I}$: ScienceQA-IMG  \cite{lu2022learn}; VQA$^\text{T}$: TextVQA  \cite{singh2019towards}; POPE  \cite{li2023evaluating}; MME  \cite{fu2023mme}; MMB$^\text{dev}$: MMBench-dev  \cite{liu2023mmbench}; Column \textit{Res.} is the image resolution of vision model. Columns \textit{PT} and \textit{IT} are the data sizes in the pre-training stage and the visual instruction tuning stage, respectively.
}
% \vspace{-6mm}
\label{tab:compare-with-sotas-vlms}
\end{table*}

 %{\color{red}{@weifei}}
\textbf{VLM training.} 
Similar to  \cite{liu2023llava,ye2023mplug}, the whole training process comprises two steps: pre-training and instruction tuning. This is depicted in Figure~\ref{fig:vlm-training-strategy}. During the first step, we freeze the vision encoder and LLM, focusing on training the efficient projector only. Subsequently, we fine-tune both the projector and LLM to enhance the abilities of visual understanding and expression by refining the language model via a language modeling loss function. Following Vicuna's hyperparameters  \cite{vicuna}, we pre-train our model on the filtered CC-595K subset  \cite{liu2023llava} for 1 epoch at a learning rate of %1e-3
$ 10^{-3} $
and a batch size of 256. We fine-tune it on the LLaVA-Instruct-158K dataset  \cite{liu2023llava} for 1 epoch at a learning rate of %2e-5
$ 2 \times 10^{-5} $
and a batch size of 128. Examples of our training dataset are shown in Figure~\ref{fig:mobile-llava-examples} (Appendix \ref{sec:VLM-demos}). We choose the AdamW optimizer with no weight decay and a cosine learning rate with a warmup ratio of 3\%. The training takes 5 hours with 8 NVIDIA Tesla A100 GPUs for \ours 1.7B, and 8 hours for \ours 3B.

\subsection{Evaluation of \ourllm}
In Table \ref{tab:llm-comparison}, we extensively assess our models on two standard natural language benchmarks, for language understanding and common sense reasoning respectively. We apply the Language Model Evaluation Harness  \cite{eval-harness} tool for the former assessment. Experimental results show that our \ourllm 1.4B is on par with the most recent open-source models such as TinyLLaMA 1.1B, Galactica 1.3B, OPT 1.3B, and Pythia 1.4B. Notably, our \ourllm 1.4B outperforms TinyLLaMA 1.1B which is trained on 2T tokens, twice as many as ours. At the 3B level, our \ourllm 2.7B also demonstrates competitive performance to INCITE 3B (V1) and OpenLLaMA 3B (V1), while \ourllm 2.7B being about 40\% faster than OpenLLaMA 3B on a Snapdragon 888 CPU as shown in Table~\ref{tab:llm-latency}.
 
For common sense reasoning, we report the zero-shot accuracy on five prevalent benchmarks, \ie, ARC$_e$  \cite{boratko2018systematic}, ARC$_c$  \cite{clark2018think}, BoolQ  \cite{clark2019boolq}, Winogrande  \cite{sakaguchi2021winogrande}, and TruthfulQA  \cite{lin2021truthfulqa}. Our models demonstrate strong reasoning capabilities on these benchmarks. It's worth noting that \ourllm 1.4B and 2.7B achieve the highest performance on Winogrande at both 1B and 3B levels. This indicates that our models deliver robust common sense capabilities, rather than cunningly relying on systematic biases in the datasets.

Further, we evaluate our models on several language understanding tasks, including PIQA  \cite{bisk2020piqa}, HellaSwag  \cite{zellers2019hellaswag}, and MMLU  \cite{hendrycks2020measuring}. We report the zero-shot accuracy on PIQA and HellaSwag, and 
 5-shot performance on MMLU. We can see that our \ourllm outperforms other models across nearly all of the benchmarks. The superior language understanding capability makes our models more suitable for downstream tasks, particularly for instruction tuning, in-context learning, etc.

%%%% MOVED HERE FOR BETTER LAYOUT AND TYPESETTING

\begin{table*}[ht]
  \centering
  \setlength{\tabcolsep}{3pt}
  \begin{tabular}{l*{8}{c}}
    \toprule
    Model & Hardware & Precision & \makecell[c]{Size \\(GB)} & \makecell[c]{$Sample$\\(tokens/s)} & \makecell[c]{$Eval_{prompt}$\\(tokens/s)} & \makecell[c]{$Eval$\\(tokens/s)} & $Total$ (s) \\
    \midrule
    \multirow{2}{*}{OpenLLaMA 3B} 
    & Snapdragon 888 & 8-bit & 3.4 & 3093 & 7.32 & 6.58 & 63.33\\
    & Snapdragon 888 & 4-bit & 2.3 & 3604 & 8.97 & 7.14 & 58.04\\

    \multirow{2}{*}{\ourllm 2.7B} 
    & Snapdragon 888 & 8-bit & 2.7 & 3919 & 17.59 & \textbf{9.14} & \textbf{44.85}\\
    & Snapdragon 888 & 4-bit & 1.5 & 3932 & 18.10 & \textbf{14.71} & \textbf{28.30} \\
    
    \midrule

    \multirow{2}{*}{TinyLLaMA 1.1B} 
    & Snapdragon 888 & 8-bit & 1.1 & 4215 & 39.49 & 19.75 & 20.83\\
    & Snapdragon 888 & 4-bit & 0.6 & 3887 & 44.17 & 31.54 & 13.22 \\

    \multirow{2}{*}{\ourllm 1.4B} 
    & Snapdragon 888 & 8-bit & 1.4  & 3846 & 35.46  & 17.93  & 22.81\\ 
    & Snapdragon 888 & 4-bit & 0.7 & 3870 & 36.20 & 28.32  & 14.76 \\

    \midrule
    \midrule
    
    \multirow{2}{*}{OpenLLaMA 3B} 
    & Jetson Orin & 8-bit & 3.4 & 2382 & 80.34 & 29.97 & 13.94\\
    & Jetson Orin & 4-bit & 2.3 & 3340 & 143.25 & 32.16 & 12.83\\
    
    \multirow{2}{*}{\ourllm 2.7B} 
    & Jetson Orin & 8-bit & 2.7 & 3040 & 133.41 & \textbf{33.28} & \textbf{12.46} \\
    & Jetson Orin & 4-bit & 1.5 & 2647 & 130.97 & \textbf{38.99} & \textbf{10.74} \\

    \midrule

    \multirow{2}{*}{TinyLLaMA 1.1B} 
    & Jetson Orin & 8-bit & 1.1 & 3007 & 279.61 & 72.30 & 5.89 \\
    & Jetson Orin & 4-bit & 0.6 & 3801 & 306.76 & 78.83 & 5.38 \\

    \multirow{2}{*}{\ourllm 1.4B} 
    &  Jetson Orin & 8-bit & 1.4 & 3289 & 249.56 & 60.73  & 6.96 \\
    &  Jetson Orin & 4-bit & 0.7 & 3738 & 253.22& 66.79  & 6.33 \\

    \bottomrule
  \end{tabular}
  \caption{\textbf{Lantency comparison of small language models on mobile and IoT devices.} ``8-bit": quantized with mode \qeight in \lammacpp and 4-bit with mode \qksn. \textit{Size} refers to the size of quantized models. $Sample$, $Eval_{prompt}$, and $Eval$ are measured in \textit{tokens per second}. $Sample$ reflects the velocity at which the next probable output token is selected, $Eval_{prompt}$ denotes the duration required to process the prompt before initiating text generation, and $Eval$ signifies the generation speed of the output tokens. $Total$ refers to the entire time consumed by a single inference (loading time included.) }
  \label{tab:llm-latency}
\end{table*}

\subsection{Comparison with SOTA VLMs}
%{\color{red}{@weifei, need analysis in depth}}
We evaluate the multimodal performance following LLaVA on  GQA  \cite{hudson2019gqa}, ScienceQA  \cite{lu2022learn},  TextVQA  \cite{singh2019towards},  POPE  \cite{li2023evaluating}, and  MME  \cite{fu2023mme}. In addition, we also exploit MMBench  \cite{liu2023mmbench} for a  comprehensive comparison.
As demonstrated in Table~\ref{tab:compare-with-sotas-vlms}, our proposed MobileVLM, despite its reduced parameters and limited training data, \textit{achieves competitive performance.} In certain instances, it even obtains superior metrics compared with the previous state-of-the-art multimodal vision language models.

In particular, on the evaluation benchmarks of GQA, POPE, and MMBench, \ours demonstrates parity with or superiority over the majority of 7B or larger VLMs, which proves its exceptional capabilities in image content perception, spatial and relational reasoning, and attribute understanding. Nonetheless, a notable shortfall arises due to the absence of large training datasets, such as code and textbook, which results in a discernible performance discrepancy on tasks like ScienceQA and MME. Still, there exists potential for enhancement in the model's proficiency by reconciling text comprehension with the nuanced recognition of fine image content on TextVQA. We list visualized demonstrations in Appendix~\ref{sec:VLM-demos}.

% To evaluate the zero-shot and few-shot learning capabilities of our models, we utilized several publicly available authoritative language model evaluation tools with a series of datasets. We compare \ourllm with the most recent open-source models (with parameter levels in the range of 1B to 3B), including 
% LLaMA (Touvron et al., 2023a),
% LLAMA 2 (Touvron et al.,
% 2023b), MPT (Mosaic ML, 2023),
% Falcon (Almazrouei et al., 2023), 
% Baichuan2 (Yang et al., 2023),
% ChatGLM2 (ChatGLM2 Team, 2023), 
% InternLM (InternLM Team, 2023), XVERSE (Inc., 2023b),
% and StableBeluga2 (Stability AI, 2023). 
% Our evaluation covers a total of several popular benchmarks, including ARC$_e$  \cite{boratko2018systematic}, MMLU  \cite{hendrycks2020measuring}, BoolQ\cite{clark2019boolq}, TruthfulQA\cite{lin2021truthfulqa}, WinoGrande  \cite{sakaguchi2021winogrande}, PIQA  \cite{bisk2020piqa}, HellaSwag  \cite{zellers2019hellaswag}, We used the Language Model Evaluation Harness  \cite{eval-harness} tool to evaluate our models.
% For the SFT model performance, we demonstrate the cases of human dialogue analysis in Appendix~\ref{sec:sft-prompt-example}.

% \subsection{multimodality Distillation}
% {\color{red}{@huyiming}}

\subsection{\ours with LoRA}
Low-Rank Adaptation (LoRA)  \cite{hu2021lora} can perform on par with or better than fully fine-tuned LLMs with fewer trainable parameters. We empirically investigate this practice on \ours to validate its multimodal performance. Specifically, during the VLM visual instruction tuning stage, we freeze all the LLM parameters except the LoRA matrices. The updated parameters are only 8.87\% and 7.41\% of the full LLM for \ourllm 1.4B and \ourllm 2.7B respectively. For LoRA configuration, we set $lora_r$ as 128 and the $lora_{\alpha}$ as 256. The results are shown in Table~\ref{tab:compare-with-sotas-vlms}. We can see that \ours with LoRA achieves comparable performances to those of full finetuning on 6 benchmarks, which is consistent with \cite{hu2021lora}.

%% MOVED HERE FOR TYPESETTING
\begin{table*}[ht]
  \centering
  \setlength{\tabcolsep}{2pt}
  \begin{tabular}{l*{9}{l}}
    \toprule
    Model & LM & Hardware & \makecell[c]{Size \\ (GB)} &  \makecell[c]{$VE$\\(ms/patch)} & \makecell[c]{$Sample$\\(tokens/s)} & \makecell[c]{$Eval_{prompt}$\\(tokens/s)} & \makecell[c]{$Eval$\\(tokens/s)} & $Total$ (s) \\
    \midrule
    \multirow{3}{*}{\makecell[c]{LLaVA-v1.5-336}}
            & Vicuna 7B & Snapdragon & 4.70 & 8.23 & 17347 & 5.36& 0.25 & 329.89  \\
            & OpenLLaMA 3B & Snapdragon & 2.88 & 7.98 & 27530 & 8.95& 7.22 & 84.43 \\
            & TinyLLaMA 1B & Snapdragon & 1.18 & 7.77 & 31370& 41.70 & 18.40& 20.70 \\
    \multirow{2}{*}{\makecell[c]{\ours-336}}
            & \ourllm 2.7B & Snapdragon & 2.14 & 8.43 & 27660 & 18.36 & \textbf{12.21} & \textbf{33.10} \\
            & \ourllm 1.4B  & Snapdragon & 1.40 & 6.82 & 34892 & 34.93 & \textbf{21.54} & \textbf{18.51} \\
    \midrule
    \multirow{3}{*}{\makecell[c]{LLaVA-v1.5-336}}
            & Vicuna 7B & Jetson Orin & 4.70 & 2.89 & 9281 & 367.26& 17.74 & 19.75  \\
            & OpenLLaMA 3B & Jetson Orin & 2.88 & 2.94 & 22270 & 474.49& 30.66 & 12.52 \\
            & TinyLLaMA 1B & Jetson Orin & 1.18 & 2.98 & 24655& 1253.94 & 76.63& 5.90 \\
    \multirow{2}{*}{\makecell[c]{\ours-336}}
            & \ourllm 2.7B & Jetson Orin & 2.14 & 3.11 & 15678 & 440.60 & \textbf{38.34} & \textbf{8.31} \\
            & \ourllm 1.4B  & Jetson Orin & 1.40 & 3.32 & 17712 & 667.69 & \textbf{65.27} & \textbf{5.14} \\
    %\multirow{\makecell{}}{}{}
    \bottomrule
  \end{tabular}
  \caption{\textbf{Lantency comparison of mobile-scale VLMs.} The language model of VLMs is quantized to 4-bit with \lammacppn. \textit{Size} is the summation of the size of the language model and the visual encoder. \textit{VE} indicates visual encoder, whose latency is measured in \textit{ms per image patch}. The remaining columns are consistent with those in Table \ref{tab:llm-latency}. LLaVA-v1.5-336-Vicuna 7B  generates 51 output tokens, while the rest VLMs generate 256 output tokens. }
  \label{tab:mobile-vlm-latency}
\end{table*}

\subsection{Latency Measurement on Mobile Devices} 
We evaluate the inference latency of MobileLLaMA and MobileVLM both on a Realme GT mobile phone and an NVIDIA Jetson AGX Orin platform. The mobile phone is equipped with a Snapdragon 888 SoC and 8GB RAM, which gives out 26 TOPS computing power. Orin is equipped with 32GB of memory, offering an impressive 275 TOPS. It operates on CUDA version 11.4, which supports the latest parallel computing technology for enhanced performance. We exploit \lammacpp \cite{Gerganov2019}  as the inference framework for both devices.

\textbf{\ourllm} %The inference speed is vital since our work targets the deployment of VLM on mobile devices. 
 For language models (LMs) in Table \ref{tab:llm-latency}, input prompts are fixed with the sentence ``\texttt{Building a website can be done in 10 simple steps:\symbol{92}nStep 1:}'', and the number of output tokens is set to 400. LMs are quantized to 4-bit and 8-bit with the quantization mode \qks and \qeight of \lammacpp, respectively. 

%First, we measure the inference speed of the language model of \ourllm on a mobile device in comparison to other small language models. Specifically, we exploit \lammacpp  \cite{Gerganov2019} framework on a Realme GT phone equipped with a Snapdragon 888 SoC and 8GB RAM for mobile inference. We ensure all of the models receive the same sentence as prompts, and generate an equal number of 400 output tokens. 
%The \textit{Mem} and \textit{CPU} are peak values caught from ``Top" command line while infering. 
%The columns of \textit{Sample}, \textit{Prompt eval}, and \textit{Eval} in Table \ref{tab:llm-latency} are measured by \lammacpp in units of \textit{tokens per second}. The load time indicates to the time it takes to load the language model, while total time refers to the time consumed for the entire inference process.

According to the measurement on the Snapdragon 888 SoC shown in Table~\ref{tab:llm-latency}, our \ourllm 2.7B significantly outperforms OpenLLaMA 3B at the same quantization precision. Notably, using two-thirds consumption of RAM and half the loading time, \ourllm 2.7B achieves \textbf{twice} the inference speed of OpenLLaMA 3B. For the smaller model, although the quantized \ourllm 1.4B is 23\% larger than that of TinyLLaMA 1B, its inference speed is only 11\% slower than TinyLLaMA 1B in terms of both \textit{Eval} speed and \textit{Total} inference time, which demonstrates that our model benefits from better architectural design. On the Orin platform, we can draw the same conclusion that {\ourllm} is quite competitive to the open-source language models at similar parameter scales.

\textbf{\ours}
Due to the limitation of \lammacpp, VLMs are split into the vision encoder and the language model, which are loaded separately during the inference stage. For the inference precision, the language model is quantized to 4-bit, while the vision encoder and the projector keep the original precision. The prompts of VLMs are composed of textual prompts and visual prompts. For a fair comparison, LLaVA-v1.5 and MobileVLMs take the picture with the same resolution of $336\times336$ as the visual prompt, along with the sentence “\texttt{What is in the picture?}” as the textual prompt. Meanwhile, the patch sizes of their vision encoder are fixed to 14. For LLaVA-v1.5, the number of input tokens (denoted as $Tks_{in}$) for the language model is composed of $576$ visual tokens and some textual tokens. For MobileVLMs, the number of visual tokens has dropped to $144$ due to the design of an efficient projector as in Sec.~\ref{Efficient Projector}. For all the tested VLMs, the number of output tokens (denoted as $Tks_{out}$) is set to 256. The total inference time of VLMs is calculated specifically as,

\begin{equation}
\begin{aligned}
Total &= Load_{LM}+(Tks_{in} / Eval_{prompt}) \\
&+ (Tks_{out}/ Sample) + (Tks_{out}/ Eval) + Others
\end{aligned}
\label{MobileVLM_Total}
\end{equation}
where $Sample$, $Eval_{prompt}$, and $Eval$ are measured in tokens per second. $Sample$ is the time it takes to ``tokenize" (sample) the prompt message. $Eval_{prompt}$ denotes the time LMs take to handle the processed tokens. $Eval$ is the time needed to generate all response tokens, 
 measured only when the LM starts emitting tokens. $Load_{LM}$ and $Total$ refer to the time of loading model and the entire time consumed by a single inference respectively. Other overheads are mostly negligible.

The inference latency comparison of various VLMs is shown in Table~\ref{tab:mobile-vlm-latency}. It can be found that $Eval$ of MobileVLM-336 is slightly slower than MobileLLaMA in Table~\ref{tab:llm-latency} since extra vision tokens increase the time consumption and consequently slow down the generation of output tokens. 

Most importantly, we are surprised to find $Total$ of MobileVLM equipped with MobileLLaMa 2.7B and MobileLLaMa 1.4B outperform LLaVA-v1.5 consistently on Snapdragon SoC and Jetson Orin. It is worth noting that although $Eval$ of MobileVLM (MobileLLaMa 1.4B) is slower than that of LLaVA-v1.5 (OpenLLaMA 1B) on Orin, it achieves a faster $Total$ inference time. The main reason is that $Tks_{in}$ of MobileVLM is only a quarter of LLaVA-v1.5 in Equation \ref{MobileVLM_Total} while other variables are comparable. This precisely proves the importance of efficient projection that reduces the number of input visual tokens.

% \begin{table*}[ht!]
%   \centering
%   \setlength{\tabcolsep}{2pt}
%   \begin{tabular}{l*{9}{c}}
%     \toprule
%     Method & LM & Vision Backbone & GQA & ScienceQA & TextVQA &  POPE & MME & MMBench \\
%     \midrule
%     LLaVA & OpenLLaMA-3B  &    &    &    &    &    &    &    \\
%     LLaVA & TinyLLaMA-1B   &    &    &    &    &    &    &    \\
%     Mobile LLaVA (Ours) & \ourllm-1.4B  &    &    &    &    &    &    &     \\
%     \bottomrule
%   \end{tabular}
%   \caption{Comparison on mainstream benchmarks with state-of-the-art VLMs.}
%   \label{tab:benchmark-on-mmlu}
% \end{table*}

% \begin{table*}[ht!]
%   \centering
%   \setlength{\tabcolsep}{2pt}
%   \begin{tabular}{l*{9}{c}}
%     \toprule
%     Setting & LM  & GQA & ScienceQA & TextVQA &  POPE & MME & MMBench \\
%     \midrule
%     full finetuning & \ourllm-1.4B      &57.08    &54.33    &44.4    &85.66    &1168.18    &53.01    \\
%     only with LoRA& \ourllm-1.4B       &    &    &    &    &    &    \\
%     full finetuning & \ourllm-2.7B     &    &    &    &    &    &     \\
%     only with LoRA& \ourllm-2.7B     &    &    &    &    &    &     \\

%     \bottomrule
%   \end{tabular}
%   \caption{Comparison of \ours with LoRA on VLM benchmarks.}
%   \label{tab:benchmark-on-lora}
% \end{table*}

\begin{table*}[t!]
	\centering
	\setlength{\tabcolsep}{4pt}
	\begin{tabular}{*{1}{c}|*{1}{l}*{2}{l}|*{6}{c}}
		\toprule
		Language Model  & Vision Encoder & VL Projector & Tokens & GQA & SQA$^\text{I}$ & VQA$^\text{T}$ & POPE  & MME & MMB$^\text{dev}$\\
		\midrule
		\multirow{5}{*}{\makecell[c]{\\ \ourllm  \\ 1.4B}} %MobileLLaMA   
		& CLIP-B-P16-S224  \cite{radford2021learning} & MLP  \cite{liu2023improvedllava} & 196 & 55.3 & 52.8 & 38.3 & 83.2 & 1079.9 & 46.1 \\
		& CLIP-L-P14-S224  \cite{radford2021learning} & MLP  \cite{liu2023improvedllava} & 256 & 55.8 & 53.4 & 40.9 & 83.9 & 1104.4 & 50.9  \\
		& CLIP-G-P14-S224  \cite{sun2023eva}     & MLP  \cite{liu2023improvedllava} & 256& \textbf{57.5} & 54.2 & 38.0 & 84.9 & 1123.3 & \textbf{53.9}  \\
		& CLIP-L-P14-S336  \cite{radford2021learning} &MLP  \cite{liu2023improvedllava} & 576 & 56.9 & 53.6 & \textbf{43.7} & \textbf{85.7} & 1137.7 & 52.8    \\
		& CLIP-L-P14-S336  \cite{radford2021learning} & LDP~(ours) & 144 & 56.1 & \textbf{54.7} & 41.5 & 84.5 & \textbf{1196.2} & 53.2    \\
		\midrule
		% \midrule
		\multirow{5}{*}{\makecell[c]{\\ \ourllm  \\ 2.7B}}   % MobileLLaMA
		& CLIP-B-P16-S224  \cite{radford2021learning}  & MLP  \cite{liu2023improvedllava} & 196 & 57.0 & 58.4 & 43.1 & 83.8 & 1212.2  & 54.6  \\
		& CLIP-L-P16-S224  \cite{radford2021learning}  & MLP  \cite{liu2023improvedllava} & 256 & 57.9 & 58.6 & 45.3 & 85.1 & 1285.0 & 57.7  \\
		& CLIP-G-P14-S224  \cite{sun2023eva}      &MLP  \cite{liu2023improvedllava} & 256& \textbf{59.5} & 58.9 & 43.9 & 85.1 & 1275.6 & 59.5  \\
		& CLIP-L-P14-S336  \cite{radford2021learning}  & MLP  \cite{liu2023improvedllava} & 576 & 59.1 & 58.3 & 47.3 & \textbf{85.8} & \textbf{1333.1} & 57.1    \\
		&CLIP-L-P14-S336  \cite{radford2021learning}   & LDP~(ours) & 144 & 59.0 & \textbf{61.0} & \textbf{47.5} & 84.9 & 1288.9 & \textbf{59.6}    \\
		\bottomrule
	\end{tabular}
	\caption{Comparison with different vision encoder scales (B/L/G: Base, Large, Giant) and visual tokens on six benchmarks with our \ourllm 1.4B/2.7B. S224/336 indicates the input resolution. }
	\label{tab:ablation-on-vision-backbones-clip}
\end{table*}

\begin{table*}[t!]
	\centering
	\setlength{\tabcolsep}{4pt}
	\scalebox{0.91}{\begin{tabular}{*{5}{l}|*{6}{c}}
		\toprule
		Vision Encoder  & Method & Pretrain Paradigm & Pretrain Data & Tokens & GQA & SQA$^\text{I}$ & VQA$^\text{T}$ & POPE  & MME & MMB$^\text{dev}$  \\
		\midrule
		% \multirow{4}{*}{\makecell[c]{ViT-B-P16-S224}} %MobileLLaMA   
		ViT-B-P16-S224 & SAIM  \cite{qi2023exploring}  & UnSupervised &IN1K  \cite{deng2009imagenet} &196 & 45.3 & 40.9& 32.2& 76.5& 845.4& 2.1 \\
		ViT-B-P16-S224 & MAE  \cite{he2022masked} & UnSupervised & IN1K  \cite{deng2009imagenet}&196 & 50.3& 49.1& 33.4& 80.2& 931.2& 24.7  \\
		\midrule
		ViT-B-P16-S224 & ViT  \cite{dosovitskiy2020image} & Classification & IN21K  \cite{ridnik2021imagenet}& 196 & 48.3 & 50.7& 33.1& 80.3& 892.8& 34.9  \\
  	% ViT-B-P16-S224 & DeiT$3$  \cite{touvron2022deit} & Classification & - & - & - &  - & - & - & -    \\
  	% Twins-SVT-B-S224 & Twins  \cite{chu2021Twins}&Classification & 49 & 49.56 & 50.51 & 32.80 & 81.08 & 954.69 & 32.90 \\
		% Twins-PCPVT-B-S224 & Twins  \cite{chu2021Twins}&Classification & 49 & 49.13 & 51.43 & 33.12 &  80.45 & 918.47 & 32.56 \\
		Twins-SVT-L-S224 & Twins  \cite{chu2021Twins}&Classification &IN1K  \cite{deng2009imagenet}& 49 & 50.3 & 50.9 & 33.2 &  80.6 & 941.6 & 32.1 \\
		% Twins-PCPVT-L-S224 & Twins  \cite{chu2021Twins}&Classification & 49 & 49.96 & 51.69 & 33.02 &  80.69 & 878.35 & 33.33 \\
        Twins-SVT-L-S384 & Twins  \cite{chu2021Twins}&Classification &IN1K  \cite{deng2009imagenet}& 144 & 51.4 & 51.0 & 32.8 &  81.7 & 930.3 & 33.4 \\
        % Twins-PCPVT-L-S384 & Twins  \cite{chu2021Twins}&Classification & 144 & 50.7 & 51.2 & 33.7 &  81.7 & 915.8 & 33.9 \\
  	Swin-Tiny-S224 & Swin  \cite{liu2021swin}  & Classification &IN1K  \cite{deng2009imagenet}& 49 & 48.3 & 50.3 & 32.5 & 80.8&  929.0& 31.4  \\
		Swin-Base-S384 & Swin  \cite{liu2021swin}  & Classification  &IN22K  \cite{deng2009imagenet}& 144 & 53.3& 52.2& 33.5& 82.8& 1037.2& 40.3   \\
		\midrule
		Swin-Tiny-S224 & GDino  \cite{liu2023grounding}  & Grounding-Det &OGC  \cite{liu2023grounding}& 49 & 51.2& 50.5& 32.4& 81.7& 932.4& 31.8  \\
		Swin-Base-S384 & GDino  \cite{liu2023grounding}  & Grounding-Det  &COGOOR  \cite{liu2023grounding} & 144 & 54.9& 51.0& 33.8& \textbf{84.5} & 1072.3& 40.0   \\
		\midrule
		ViT-B-P14-S224 & CLIP   \cite{radford2021learning} & Image-Text-Align &WIT  \cite{radford2021learning}& 256 & \textbf{55.3} & \textbf{52.8} & \textbf{38.3} & 83.2 & \textbf{1079.9} & \textbf{46.1}  \\
		\bottomrule
	\end{tabular}}
	\caption{Comparison with various vision encoders from different pre-training paradigms on \ourllm 1.4B on six benchmarks.}
	\label{tab:ablation-on-vision-backbones-pt}
\end{table*}

\section{Ablation Study}

\subsection{Ablation on Vision Backbones}~\label{sec:ablation-vision} 
In Table \ref{tab:ablation-on-vision-backbones-clip}, we compare the multimodal performance at varying model scales and different numbers of visual tokens. All experiments are conducted with CLIP ViT as a visual encoder. We configure different model scales, patch sizes, and types of vision-language projectors.

\textbf{The impact of model scales.}
As the model scales up, the multimodal performance on 6 benchmarks maintains a gradual increase trend under the same projector  \cite{liu2023improvedllava}. 
However, it can be observed that the gain brought by the visual model scaling may gradually become saturated at a certain amount of training data.

\textbf{The impact of the number of visual tokens.}
Compared with rows 4-5, our proposed LDP module reduces the number of visual tokens from 576 to 144 ($\downarrow$ 75\%), and it finally achieves performance equivalent to or sometimes better than the baseline. This reveals that the quality of visual tokens can be further improved while our proposed LDP module is quite effective.

\textbf{The impact of pre-training paradigms.}  Furthermore, we show the performance of \ourllm 1.4B under different vision backbone pre-training paradigms in Table~\ref{tab:ablation-on-vision-backbones-pt}. Based on the cost of annotation and pre-training, we roughly classify these paradigms into four categories. 
%, \ie, unsupervised training, supervised classification training, supervised detection training, and supervised image-text-align training. 
It turns out that the performance of \ours gradually improves as the pre-training cost increases. The vision encoder pre-trained with supervised image-text alignment achieved the best performance.
By comparing \textit{Swin-Base-S384-GDino} and \textit{ViT-B-P14-S224}, we notice that the model pre-trained by grounding detection achieved relatively comparable performance to the CLIP pre-trained model on GQA, SQA, POPE, and MME. This outcome indicates that the \textit{\textbf{image-level}} alignment has greater potential to strike better performance than  \textbf{\textit{object-level}}, especially by using more visual tokens or more training data.
In addition, better ImageNet performance of pre-trained models (\eg, Swin $>$ ViT) often corresponds to more general visual feature extraction capabilities, and \ours will have certain performance gains in turn.

\begin{table*}[t!]
	\centering
	\setlength{\tabcolsep}{4pt}
	\begin{tabular}{*{1}{l}|*{1}{c}|*{6}{c}}
		\toprule
		VL Projector Architecture Design  & Tokens  & GQA & SQA$^\text{I}$ & VQA$^\text{T}$ & POPE  & MME & MMB$^\text{dev}$  \\
		\midrule
		% \multirow{4}{*}{\makecell[c]{ViT-B-P16-S224}} %MobileLLaMA
		\rowcolor{timberwolf}$[{PW}]_{\times 2}[DW^{\kappa = 1}PW]_{\times 0}[DW^{\kappa = 2}PW]_{\times 0}$  & 576  & 56.9 & 53.6 & 43.7 & 85.7 & 1137.7 & 52.8\\
		% $[{PW}]_{\times 0} \succ [DW^{\kappa = 1}PW]_{\times 2} \succ [DW^{\kappa = 2}PW]_{\times 0}$  & 576   & - & - & - & - & - & - \\
		$[{PW}]_{\times 0}[DW^{\kappa = 1}PW]_{\times 1}[DW^{\kappa = 2}PW]_{\times 1}$  & 144   & 54.9 &52.9& 40.2& 84.0& 1150.8& 50.3 \\
		\rowcolor{mygray} $[{PW}]_{\times 2}[DW^{\kappa = 1}PW]_{\times 1}[DW^{\kappa = 2}PW]_{\times 1}$  & 144   &56.1 & 54.7 & 41.5 & 84.5 & 1196.2 & 53.2\\
		$[{PW}]_{\times 2}[DW^{\kappa = 1}PW]_{\times 3}[DW^{\kappa = 2}PW]_{\times 1}$ & 144  & 55.3& 53.9& 40.8& 84.6&1166.3& 53.0  \\		
		$[{PW}]_{\times 2}[DW^{\kappa = 2}PW]_{\times 1}[DW^{\kappa = 1}PW]_{\times 1}$  & 144  & 55.6 & 54.3 & 41.5& 84.6& 1166.2& 52.8\\
		\bottomrule
	\end{tabular}
	\caption{The exploration of projector design based on \ourllm 1.4B. The $PW$ represents $pointwise$-$conv$ and $DW$ is $depthwise$-$conv$. The subscript $\times$ indicates the number of times the corresponding module is stacked repeatedly. The superscript $\kappa$ indicates the $conv$ stride.
    %of $depthwise$-$conv$.
    The \colorbox{timberwolf}{grey row} is the baseline projector from \cite{liu2023improved}, and \colorbox{mygray}{green row} is the proposed LDP in our \ours.}
	\label{tab:ablation-on-vision-backbones-arch}
\end{table*}

%%% MOVED HERE FOR BETTER TYPESETTING

\begin{table*}[t!]
\centering
\setlength{\tabcolsep}{4pt}
\begin{tabular}{*{3}{c}|*{6}{c}}
\toprule
LLM base model & SFT strategy & Conversation mode & GQA & SQA$^\text{I}$ & VQA$^\text{T}$ & POPE  & MME & MMB$^\text{dev}$\\
\midrule
\ourllm 1.4B      & w/o  & llava$_{llama_2}$ & 55.8 & 52.6 & 40.5 &  84.4& 1111.5 & 52.0\\
\ourllm 1.4B      & w/o  & vicuna$_{v1}$ & 56.1 & 53.0 & 40.5& 84.6& 1148.5& 50.3\\
\ourllm 1.4B      & Alpaca  & llava$_{llama_2}$  &  55.2 & 54.8 & 40.6 & 84.9 & 1171.1 & 51.9\\
\ourllm 1.4B      & Alpaca  & vicuna$_{v1}$  & 55.5  & 53.1 & 40.6 & 83.8 & 1168.0 & 47.7\\
\rowcolor{mygray} \ourllm 1.4B      & Vicuna &  vicuna$_{v1}$  &56.1 & 54.7 & 41.5 & 84.5 & 1196.2 & 53.2\\
\bottomrule
\end{tabular}
\caption{Quantitative analysis on SFT of MobileLLaMA 1.4B in downstream tasks.}
\label{tab:quantitive_of_SFT}
\end{table*}

\subsection{Abaltion on VL Projectors}
Motivated by the fact both feature interaction and token interaction are beneficial, we utilize depthwise convolutions for the former and pointwise for the latter. Table \ref{tab:ablation-on-vision-backbones-arch} shows the performance of various VL projectors.
Row $1$ in Table \ref{tab:ablation-on-vision-backbones-arch} is the module used in LLaVA  \cite{liu2023llavaplus} where only the feature space is transformed through two $linear$ layers. 
Row $2$ adds a $DW$ (depthwise) convolution before each $PW$(pointwise) for token interaction, which performs $2\times$ downsampling with a stride of $2$.  We notice that the performance begins to show an evident decline.
Based on the setup of $144$ tokens, adding two front-end $PW$ layers brings more feature-level interactions, which makes up for the performance loss caused by token reduction.
Rows $4$ and $5$ show that adding more parameters does not achieve desired gains. Rows $4$ and $6$ show that the downsampling of tokens at the end of the projector has a positive effect.

\subsection{Visual Resolution and Token Numbers}
Since the number of visual tokens directly affects the inference speed of the whole multimodal model, we compare two types of design: reducing the input resolution (RIR) and using a lightweight downsample projector (LDP). Without loss of generality, for an image of $H \times W$ with a patch size of $P$, the former strategy generates $HW/P^2$ tokens. For the latter, it produces $HW/4P^2$ tokens using a downsampling ratio of 2. We use $H=W=336, P=14$ for LDP  and $H=W=168, P=14$ for RIR to keep the total number of tokens as 144. The result from   Table~\ref{tab:quantitive_of_token_reductioon} verifies the effectiveness of the proposed LDP.
\begin{table}[h]
\centering
\setlength{\tabcolsep}{3pt}
\begin{tabular}{*{1}{l}|*{6}{c}}
\toprule
Method  & GQA & SQA$^\text{I}$ & VQA$^\text{T}$ & POPE  & MME & MMB$^\text{dev}$\\
\midrule
LDP  & 56.1 & 54.7 & 41.5 & 84.5 & 1196.2 & 53.2 \\
RIR & 53.9  & 53.1  & 37.1 & 81.5 & 1072.5 & 46.7 \\
%\rowcolor{mygray}
\bottomrule
\end{tabular}
\caption{Token reduction design on \ours{ 1.7B}.}
\label{tab:quantitive_of_token_reductioon}
\end{table}

\subsection{Quantitative Analysis on SFT}~\label{sec:exp-sft}
Vicuna  \cite{zheng2023judging}, fine-tuned on LLaMA, has been widely chosen in large multimodal models  \cite{liu2023improvedllava, dai2023instructblip, zhu2023minigpt}. We further explore how  SFT affects our language model's performance in downstream tasks. Two common SFT paradigms Alpaca  \cite{alpaca} and Vicuna  \cite{zheng2023judging} are compared in Table~\ref{tab:quantitive_of_SFT}.  We find that the scores of SQA, VQA, MME, and MMBench can all be significantly enhanced. It demonstrates that fine-tuning large language models in Vicuna dialogue mode  \cite{zheng2023judging} with the data from ShareGPT ultimately achieves the best performance. To better integrate SFT's prompt format with the training of downstream tasks, we ablate the conversation mode on \ours to find vicuna$_{v1}$ performs best.

% \subsection{Broader Impact}
% We present how to build a strong and reproducible mobile vision language assistant from scratch and in greater detail. It is a step forward to enjoy the power of multimodal models in many scenarios where existing VLM models cannot cover, such as limited resources and privacy protection. Since making the whole pipeline reproducible is one of our targets, we only make use of public data. We hope this will be a good starting point for the VLM community. Cautiously note that our method also suffers from unexpected hallucinations as spotted in large language models.

\section{Conclusion}
\label{sec:concl}

In a nutshell, we present \ours, a set of efficient and high-powered mobile scale vision language models tailored for mobile and IoT devices. In its making, we refurbish both language models and vision projection modules. Extensive experiments are conducted to choose proper vision backbones, to design an efficient projector, and to enhance model capacity by training schemes like language model SFT, a two-stage training strategy involving pretraining and instruction tuning, and LoRA fine-tuning. The performance is evaluated vigorously on mainstream VLM benchmarks. MobileVLMs also show an unprecedented speed on typical mobile and IoT devices. We believe 
that 
\ours will open up new possibilities for widespread applications like multimodal assistants deployed on mobile devices or in self-driving cars, and more broadly embodied AI robots.

\paragraph{Acknowledgements:}   This work was
%The participation of X. Chu, B. Zhang and C. Shen was 
in part 
supported by National Key R\&D Program of China (No.\  2022\-ZD\-0118\-700).

The authors thank Yerui Sun for infrastructure support.

%%%%%%%%% REFERENCES
\clearpage
\clearpage
{\small
\bibliographystyle{ieee_fullname}
\bibliography{egbib}
}

%%%%%%%%%%% APPENDIX
\clearpage
\clearpage
\appendix
\section{Dataset}

\subsection{Dataset for \ourllm}\label{sec:dataset-for-ourllm}
Table~\ref{tab: LLM_traing_data_proportion} shows the sampling strategy for the training set of MobileLLaMA.

\begin{table}[ht]
  \centering
  \begin{tabular}{@{}lccccc@{}}
    \toprule
    Dataset & Proportion & \\
    \midrule
    Arxiv & 2.5\%\\
    Book & 4.5\%\\
    C4 & 15.0\%\\
    Common Crawl & 67.7\%\\
    Github & 4.5\% \\
    StackExchange & 2.0\%\\
    Wikipedia & 4.5\%\\
    \bottomrule
  \end{tabular}
  \caption{Data sample strategy for \ourllm training.}
  \label{tab: LLM_traing_data_proportion}
\end{table}

\section{SFT prompt and response example}
We present our prompt template of \ourllm SFT in Table \ref{table:sft-prompt}.
% \label{sec:sft-prompt-example}
\sloppy
\begin{table*}[h]
\centering
\begin{tabular}{p{0.24\textwidth}p{0.68\textwidth}}
    \toprule
    Human: & \texttt{who are you?}  \\
    \midrule
    \ourllm 1.4B Base: &  \texttt{I'm the AI.} \\
    \midrule
    \ourllm 1.4B SFT: &  \texttt{I am a digital asset management and metadata expert. My expertise includes creating and managing digital content, metadata, and digital asset management solutions. I have a passion for helping organizations unlock the value of their digital assets and improve their workflows. I am also experienced in working with big data and analytics and can help organizations unlock insights from their data to drive better decision-making. I am always up for a challenge and love working with clients to help them achieve their goals.} \\
    \midrule
    Human: & \texttt{what can you do?}  \\
    \midrule
    \ourllm 1.4B Base: &  \texttt{I'm a chatbot.} \\
    \midrule
    \ourllm 1.4B SFT: &  \texttt{In addition to providing expert advice and guidance on a wide range of topics, I can also help you with tasks and projects that are more complex or require specialized knowledge, such as developing and implementing project plans or managing budget and resource allocation.
    Overall, my role is to provide value and support to your project or task and to help you achieve your goals in a timely and efficient manner.} \\
    \bottomrule
\end{tabular}
\caption{Examples of the response of the \ourllm1.4B Base and SFT model}
\label{table:sft-model-humal-eval}
\end{table*}

\sloppy
\begin{table*}[h]
\centering
\begin{tabular}{p{0.1\textwidth}p{0.8\textwidth}}
    \toprule
    System & \texttt{A chat between a curious user and an artificial intelligence assistant. The assistant gives helpful, detailed, and polite answers to the user's questions.}  \\
    \midrule
    USER &  \texttt{I want you to act as my travel agent for preparing an itinerary for travel to Munnar and Tekkady in Kerala. I have already booked flights from Hyderabad to Kochi for an onward journey on 25th Jan and a return journey on 28th Jan. We are a group of 4 men and planning to stay 2 days in Munnar and 1 day in Tekkady. I want you to help me with accommodation booking and activities.} \\
    \midrule
    ASSISTANT &  \texttt{I'd be happy to help you plan your trip to Munnar and Tekkady in Kerala! Here is a suggested itinerary for your trip: \symbol{92}n\symbol{92}nDay 1: Arrival in Kochi - Travel to Munnar (around 4 hours)\symbol{92}n\symbol{92}n* Check-in at a hotel in Munnar\symbol{92}n* Afternoon visit to the tea plantations and factories\symbol{92}n* Evening at leisure\symbol{92}n\symbol{92}nDay 2: Munnar\symbol{92}n\symbol{92}n* Early morning visit to the Eravikulam National Park to see the endangered Nilgiri Tahr\symbol{92}n* Afternoon visit to the Mattupetty Dam and the Kundala Dam\symbol{92}n* Evening at leisure\symbol{92}n\symbol{92}nDay 3: Munnar to Tekkady (around 4 hours)\symbol{92}n\symbol{92}n* Check-in at a hotel in Tekkady\symbol{92}n* Afternoon visit to the Periyar Wildlife Sanctuary for a boat ride and wildlife viewing\symbol{92}n* Evening at leisure\symbol{92}n\symbol{92}nDay 4: Tekkady - Return to Kochi\symbol{92}n\symbol{92}n* Check-out and travel back to Kochi\symbol{92}n* Flight back to Hyderabad\symbol{92}n\symbol{92}nIn terms of accommodation, there are many options available in Munnar and Tekkady. You can choose from budget-friendly options to luxury resorts. You can check the availability and prices on hotel booking websites like Booking.com, Agoda, Expedia, etc.\symbol{92}n\symbol{92}nPlease let me know if you have any other questions or if you need further assistance with your trip planning.</s>} \\
    \bottomrule
\end{tabular}
\caption{An example prompt and annotation of our SFT data.}
\label{table:sft-prompt}
\end{table*}

\section{VLM demos}
\begin{figure*}[t]
  \centering
   \includegraphics[width=0.85\linewidth]{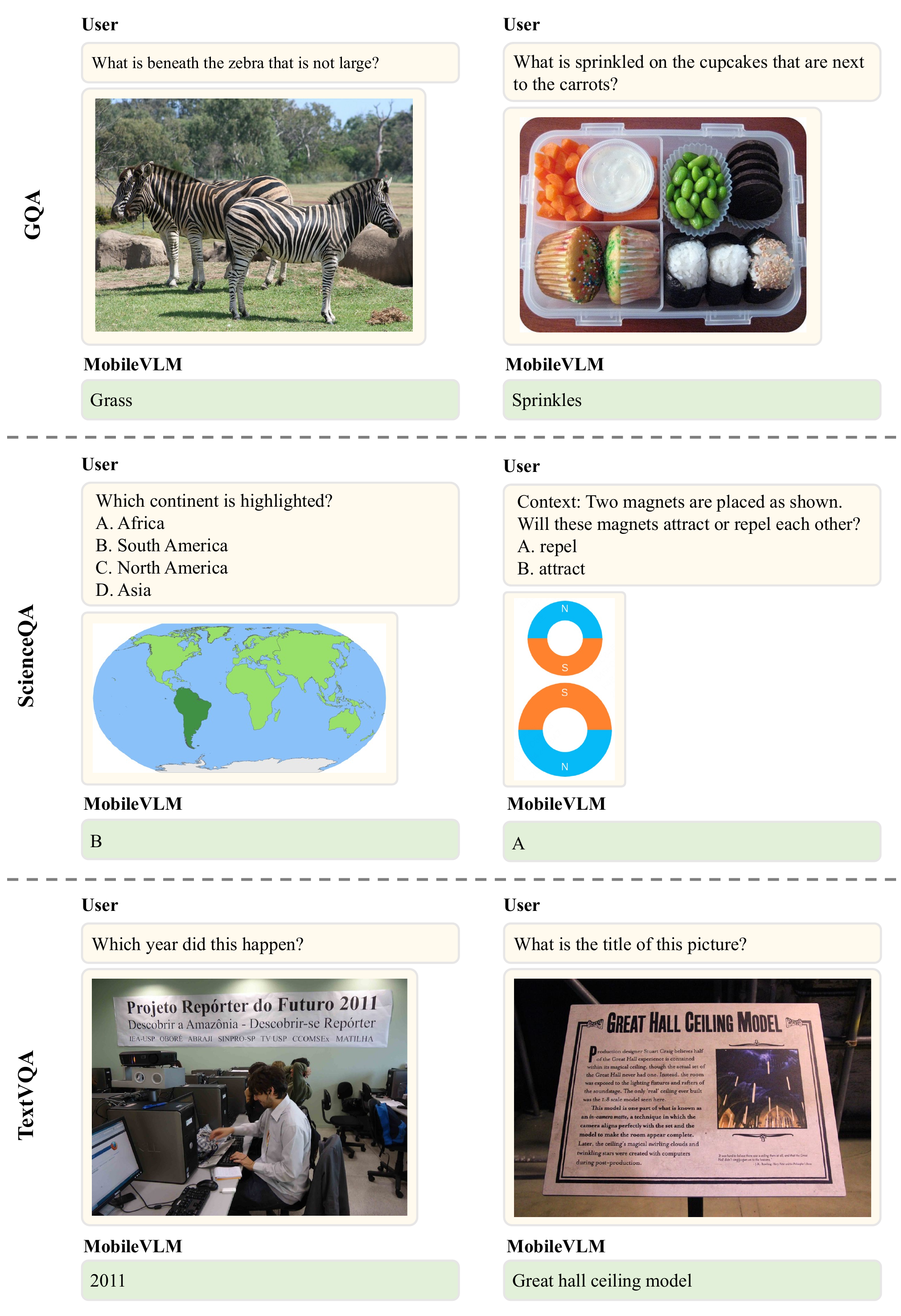}
   \caption{Examples of \ours on GQA, ScienceQA, and TextVQA Benchmarks.}
   \label{fig:mobile-llava-examples-benchmark1}
\end{figure*}

\begin{figure*}[t]
  \centering
   \includegraphics[width=0.85\linewidth]{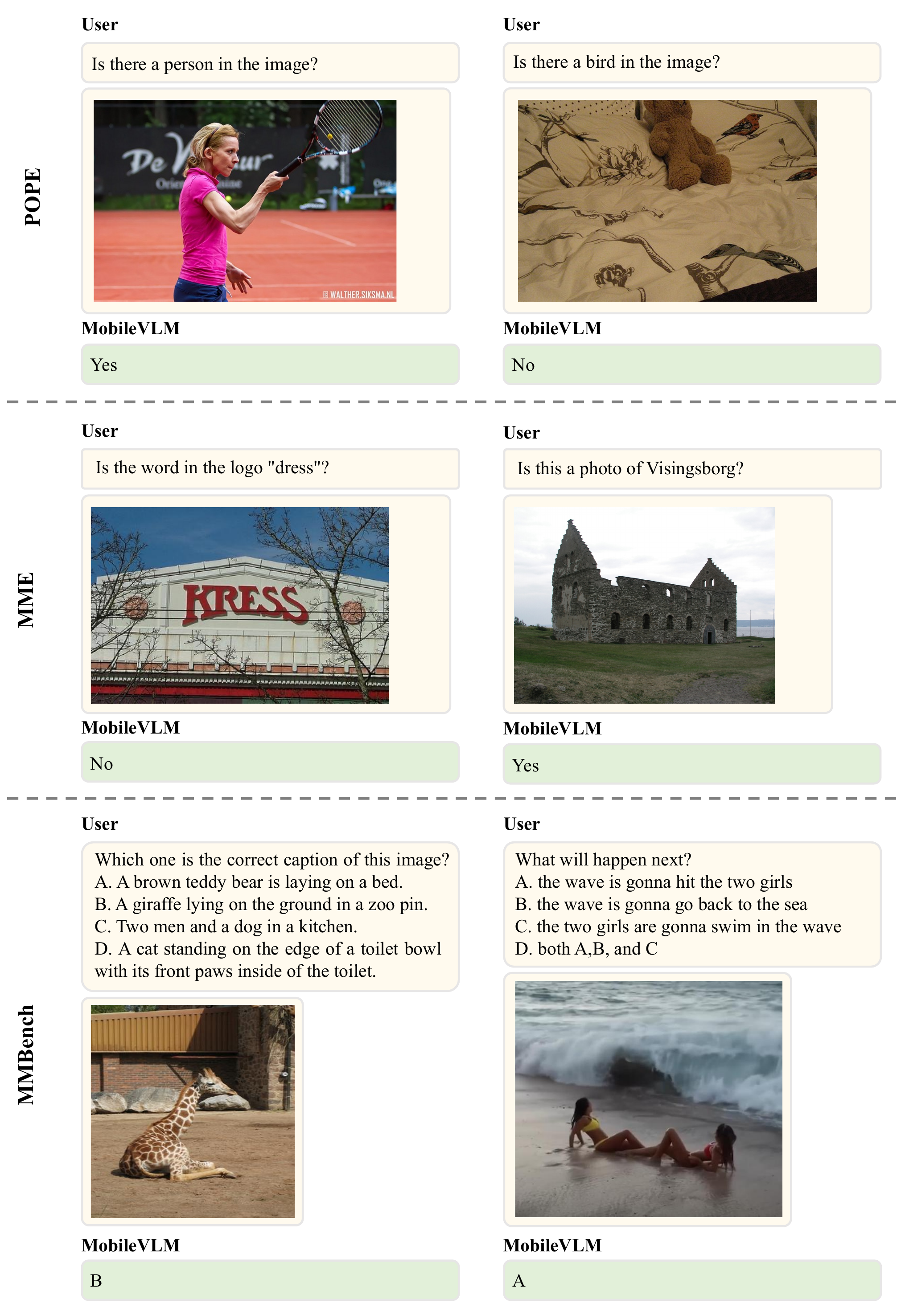}
   \caption{Examples of \ours on POPE, MME and MMBench Benchmarks.}
   \label{fig:mobile-llava-examples-benchmark2}
\end{figure*}

 Figure~\ref{fig:mobile-llava-examples-benchmark1} and Figure~\ref{fig:mobile-llava-examples-benchmark2} showcase MobileVLM's performance on each benchmark, including the scenes on attribute understanding (GQA left), spatial and relational reasoning (GQA right, MMBench), social science (ScienceQA left), natural science (ScienceQA right), OCR (TextVQA), object recognition (POPE), word knowledge (MME right), \etc.
 
\label{sec:VLM-demos}
\begin{figure*}[t]
  \centering
   \includegraphics[width=\linewidth]{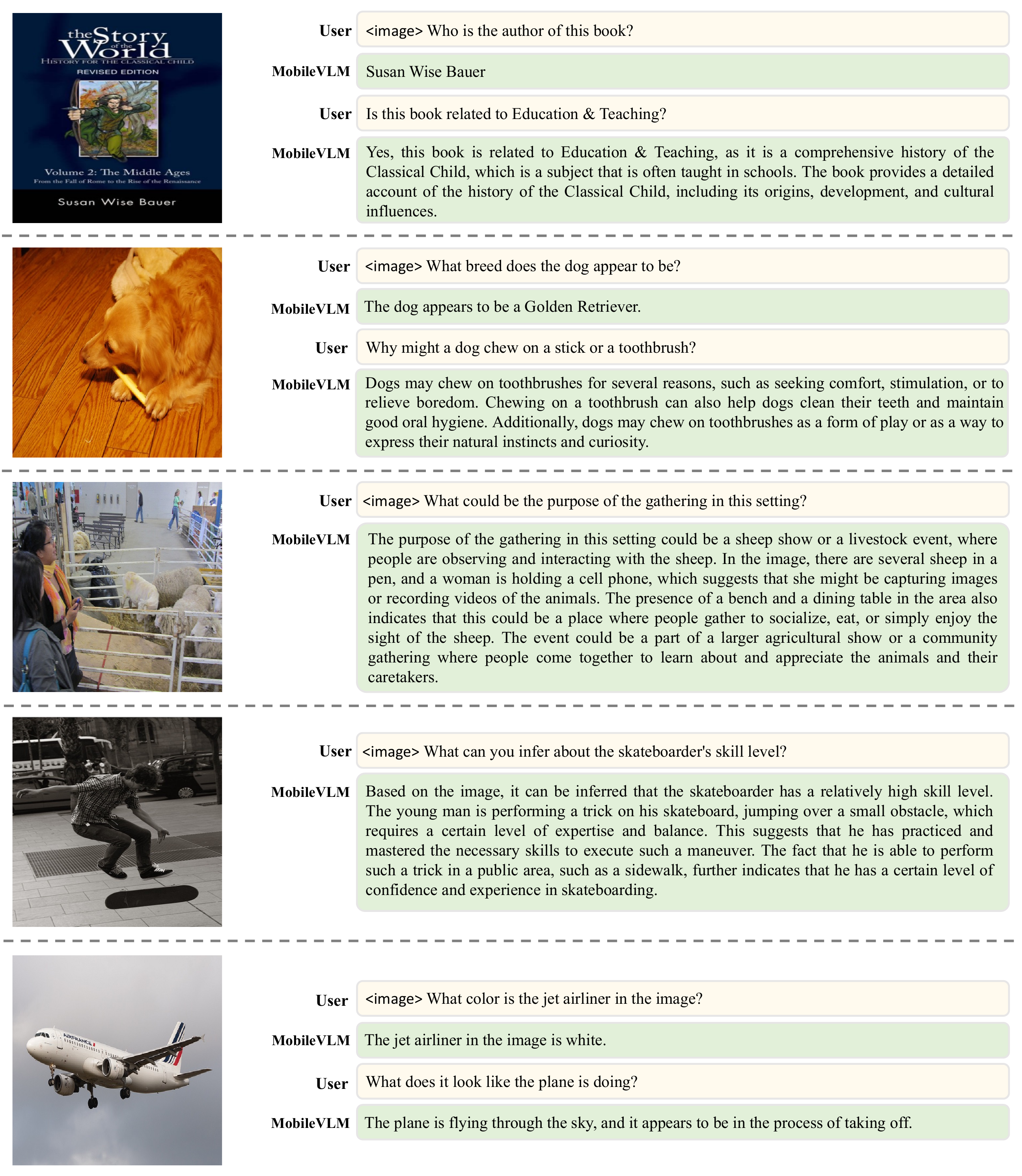}
   \caption{Examples of \ours on VLM training datasets.}
   \label{fig:mobile-llava-examples}
\end{figure*}

\end{document}